%% file: neurips_2022.tex
\documentclass{article}

\pdfoutput=1

\usepackage{graphicx}
\usepackage{amsmath}
\usepackage{amssymb}
\usepackage{booktabs}
\usepackage{wrapfig}

\usepackage{multirow}
\usepackage{verbatim}
\usepackage{enumitem}
\usepackage{bm}

\usepackage{subcaption}

\newcommand{\xv}{{\boldsymbol x}}
\newcommand{\yv}{{\boldsymbol y}}
\newcommand{\rv}{{\boldsymbol R}_{\textsc{TA}}}
\newcommand{\tv}{{\boldsymbol T}_{\textsc{IA}}}
\newcommand{\iv}{{\boldsymbol I}_{\textsc{TA}}}

\usepackage{xcolor}
\definecolor{bittersweet}{rgb}{1.0, 0.44, 0.37}
\definecolor{mygreen}{rgb}{0.29, 0.7, 0.48}

\newcommand{\tabincell}[2]{\begin{tabular}{@{}#1@{}}#2\end{tabular}}

\usepackage[font=small,labelfont=bf]{caption}  
\usepackage{makecell}
\usepackage{tabulary}
\definecolor{demphcolor}{RGB}{144,144,144}
\newcommand{\demph}[1]{\textcolor{demphcolor}{#1}}
\definecolor{mygray}{gray}{0.4}
\usepackage{pifont}

\usepackage{xspace}

\def\ModelName{\textsc{Fiber}\xspace}

\newcommand\blfootnote[1]{%
  \begingroup
  \renewcommand\thefootnote{}\footnote{#1}%
  \addtocounter{footnote}{-1}%
  \endgroup
}

\usepackage[pagebackref,breaklinks,colorlinks]{hyperref}

\usepackage[preprint,nonatbib]{neurips_2022}

\usepackage[utf8]{inputenc} 
\usepackage[T1]{fontenc}    
\usepackage{hyperref}       
\usepackage{url}            
\usepackage{booktabs}       
\usepackage{amsfonts}       
\usepackage{nicefrac}       
\usepackage{microtype}      
\usepackage{xcolor}         

\title{Coarse-to-Fine Vision-Language Pre-training with Fusion in the Backbone }

\author{
\textbf{\normalsize{Zi-Yi Dou$^{*\ddagger}$, Aishwarya Kamath$^{*\natural}$, Zhe Gan$^{*\dagger\spadesuit}$, Pengchuan Zhang$^{\mathsection}$, Jianfeng Wang$^{\dagger}$}} \\ \textbf{\normalsize{Linjie Li$^{\dagger}$, Zicheng  Liu$^{\dagger}$, Ce Liu$^{\dagger}$, Yann LeCun$^{\natural}$, Nanyun Peng$^{\ddagger}$, Jianfeng Gao$^{\dagger}$, Lijuan Wang$^{\dagger}$}} \\
\normalsize{$^{\dagger}$Microsoft}~~~~$^{\ddagger}$University of California, Los Angeles~~~~$^{\natural}$New York University \\
{\tt\small \{zdou,violetpeng\}@cs.ucla.edu, \{aish,yann.lecun\}@nyu.edu, pengchuanzhang@fb.com}  \\
{\tt\small \{zhgan,jianfw,linjli,zliu,liuce,jfgao,lijuanw\}@microsoft.com}
}

\begin{document}

\maketitle

\begin{abstract}
Vision-language (VL) pre-training has recently received considerable attention. However, most existing end-to-end pre-training approaches either only aim to tackle VL tasks such as image-text retrieval, visual question answering (VQA) and image captioning that test high-level understanding of images, or only target region-level understanding for tasks such as phrase grounding and object detection. We present \ModelName (\textbf{F}usion-\textbf{I}n-the-\textbf{B}ackbone-based transform\textbf{ER}), a new VL model architecture that can seamlessly handle both these types of tasks. Instead of having dedicated transformer layers for fusion after the uni-modal backbones, \ModelName pushes multimodal fusion deep into the model by inserting cross-attention into the image and text backbones, bringing gains in terms of memory and performance. In addition, unlike previous work that is either only pre-trained on image-text data or on fine-grained data with box-level annotations, we present a two-stage pre-training strategy that uses both these kinds of data efficiently: ($i$) \emph{coarse}-grained pre-training based on image-text data; followed by ($ii$) \emph{fine}-grained pre-training based on image-text-box data. We conduct comprehensive experiments on a wide range of VL tasks, ranging from VQA, image captioning, and retrieval, to phrase grounding, referring expression comprehension, and object detection. Using deep multimodal fusion coupled with the two-stage pre-training, \ModelName provides consistent performance improvements over strong baselines across all tasks, often outperforming methods using magnitudes more data. Code is available at \url{https://github.com/microsoft/FIBER}. \blfootnote{$^*$Equal Technical Contribution\hspace{3mm}$^\spadesuit$Project Lead \hspace{3mm}$^\mathsection$Work done while at Microsoft}
\end{abstract}

\input{subtex/1_intro}

\input{subtex/2_related_work}

\input{subtex/3_method}
\input{subtex/4_experiments}

\input{subtex/5_conclusions}

{\small
\bibliographystyle{plain}
\bibliography{reference}
}

\section*{Checklist}

\begin{enumerate}

\item For all authors...
\begin{enumerate}
  \item Do the main claims made in the abstract and introduction accurately reflect the paper's contributions and scope?
    \answerYes{}
  \item Did you describe the limitations of your work?
    \answerYes{See Section \ref{sec:limitations}}
  \item Did you discuss any potential negative societal impacts of your work?
    \answerYes{See Section \ref{sec:limitations}}
  \item Have you read the ethics review guidelines and ensured that your paper conforms to them?
    \answerYes{}
\end{enumerate}

\item If you are including theoretical results...
\begin{enumerate}
  \item Did you state the full set of assumptions of all theoretical results?
     \answerNA{}
        \item Did you include complete proofs of all theoretical results?
     \answerNA{}
\end{enumerate}

\item If you ran experiments...
\begin{enumerate}
  \item Did you include the code, data, and instructions needed to reproduce the main experimental results (either in the supplemental material or as a URL)?
    \answerYes{} 
  \item Did you specify all the training details (e.g., data splits, hyperparameters, how they were chosen)?
    \answerYes{}
        \item Did you report error bars (e.g., with respect to the random seed after running experiments multiple times)?
    \answerNo{}
        \item Did you include the total amount of compute and the type of resources used (e.g., type of GPUs, internal cluster, or cloud provider)?
    \answerYes{}
\end{enumerate}

\item If you are using existing assets (e.g., code, data, models) or curating/releasing new assets...
\begin{enumerate}
  \item If your work uses existing assets, did you cite the creators?
    \answerYes{}
  \item Did you mention the license of the assets?
    \answerYes{}
  \item Did you include any new assets either in the supplemental material or as a URL?
     \answerNA{}
  \item Did you discuss whether and how consent was obtained from people whose data you're using/curating?
     \answerNA{}
  \item Did you discuss whether the data you are using/curating contains personally identifiable information or offensive content?
     \answerNA{}
\end{enumerate}

\item If you used crowdsourcing or conducted research with human subjects...
\begin{enumerate}
  \item Did you include the full text of instructions given to participants and screenshots, if applicable?
    \answerNA{}
  \item Did you describe any potential participant risks, with links to Institutional Review Board (IRB) approvals, if applicable?
     \answerNA{}
  \item Did you include the estimated hourly wage paid to participants and the total amount spent on participant compensation?
    \answerNA{}
\end{enumerate}

\end{enumerate}

\clearpage
\input{subtex/6_supp}

\end{document}

%% file: subtex/1_intro.tex
\vspace{-3mm}
\section{Introduction}
\vspace{-2mm}
Inspired by the success of language model pre-training~\cite{devlin2018bert,raffel2020exploring,liu2019roberta}, coupled with the unification of architectures used in the NLP and computer vision communities \cite{dosovitskiy2020image, carion2020end}, vision-language pre-training (VLP)~\cite{tan-bansal-2019-lxmert,lu2019vilbert,li2019visualbert,chen2020uniter} has been receiving an increasing amount of attention. It has been proven that VLP can establish state-of-the-art performance on visual question answering~\cite{antol2015vqa}, visual reasoning~\cite{suhr2018corpus}, image captioning, and image-text retrieval~\cite{Lin2014MicrosoftCC}. The pre-training objectives commonly used for these tasks, such as image-text matching, image conditioned masked language modeling and image-text constrastive learning, require multimodal understanding at the image level. Typically, this means the pre-training is done using images at lower resolution (\emph{e.g.}, 384$\times$384), making it possible to scale up training by using large batch sizes.

Recently, it has also been shown that tasks such as image classification and object detection (OD), which have been traditionally viewed as vision-only tasks, can benefit from being cast as VL tasks~\cite{radford2021learning,jia2021scaling,li2021grounded,kamath2021mdetr}. Inspired by MDETR \cite{kamath2021mdetr}, GLIP~\cite{li2021grounded} reformulates standard classification-based OD as phrase grounding. This opens up the possibility to leverage VLP for OD, and vice versa, and this unification has led to impressive performance on several established OD as well as phrase grounding benchmarks \cite{plummer2015flickr30k}. Since these tasks involve fine-grained image understanding between regions in the image and phrases in the text, and also require prediction of precise bounding boxes at the output, the pre-training typically involves using high resolution input images (\emph{e.g.}, 800$\times$1,333).

\begin{figure*}
  \centering
    \includegraphics[width=1.0\linewidth]{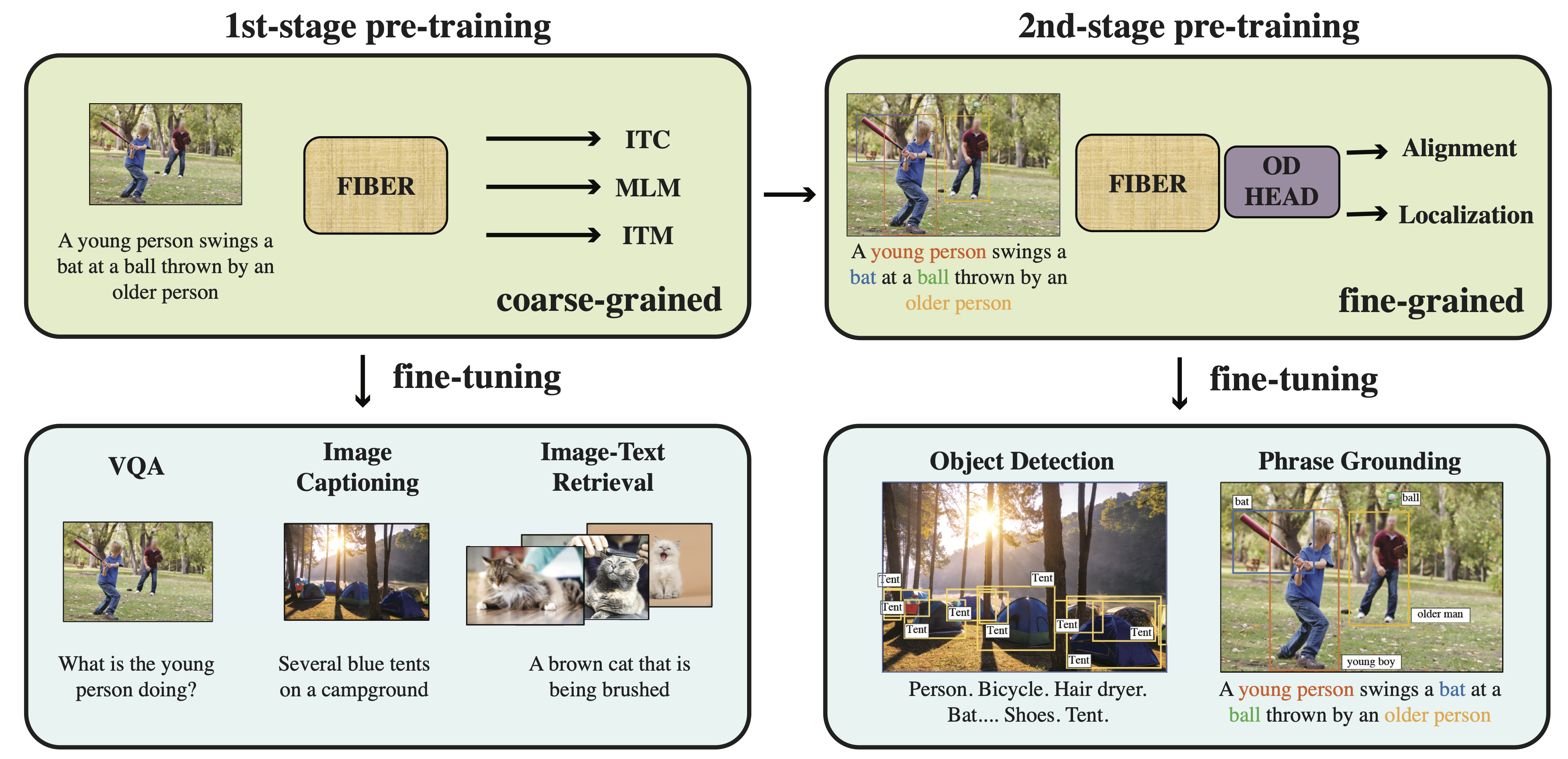}
  \caption{The proposed coarse-to-fine pre-training framework for vision-language tasks. We first perform \emph{coarse}-grained pre-training with image-text data for VQA, image captioning and retrieval tasks, and then perform \emph{fine}-grained pre-training with image-text-box data for phrase grounding and object detection tasks. The same \ModelName architecture is used for both stages. OD: object detection. MLM: masked language modeling. ITM: image-text matching. ITC: image-text contrastive loss.}
  \label{fig:main}
  \vspace{-6.5mm}
\end{figure*}

Existing multimodal architectures typically do not support both kinds of tasks. Specifically, the fully end-to-end VLP models such as ALBEF~\cite{li2021align}, METER~\cite{dou2021empirical}, and SimVLM~\cite{wang2021simvlm} can achieve the state of the art (SoTA) on image-level understanding tasks, but it is non-trivial to extend them for region-level VL tasks because predicting bounding boxes is typically hard in end-to-end settings. On the other hand, MDETR~\cite{kamath2021mdetr} and GLIP~\cite{li2021grounded} are designed to predict bounding boxes, but have not been shown to support tasks such as image captioning and retrieval. Further, fine-grained pre-training not only requires data with bounding box annotations that are cumbersome to acquire, but the requirement of high input image resolution makes pre-training very costly, especially when using standard Transformer architectures~\cite{vaswani2017attention} that have quadratic complexity in the size of the image. 
A natural but challenging question arises: \emph{can we have a unified framework for efficient VL pre-training that benefits both image-level and region-level VL tasks (e.g., both VQA and OD)?}

\begin{wrapfigure}{R}{0.5\textwidth}
\centering
\vspace{-5mm}
\includegraphics[trim=100 90 100 100,clip,width=0.5\textwidth]{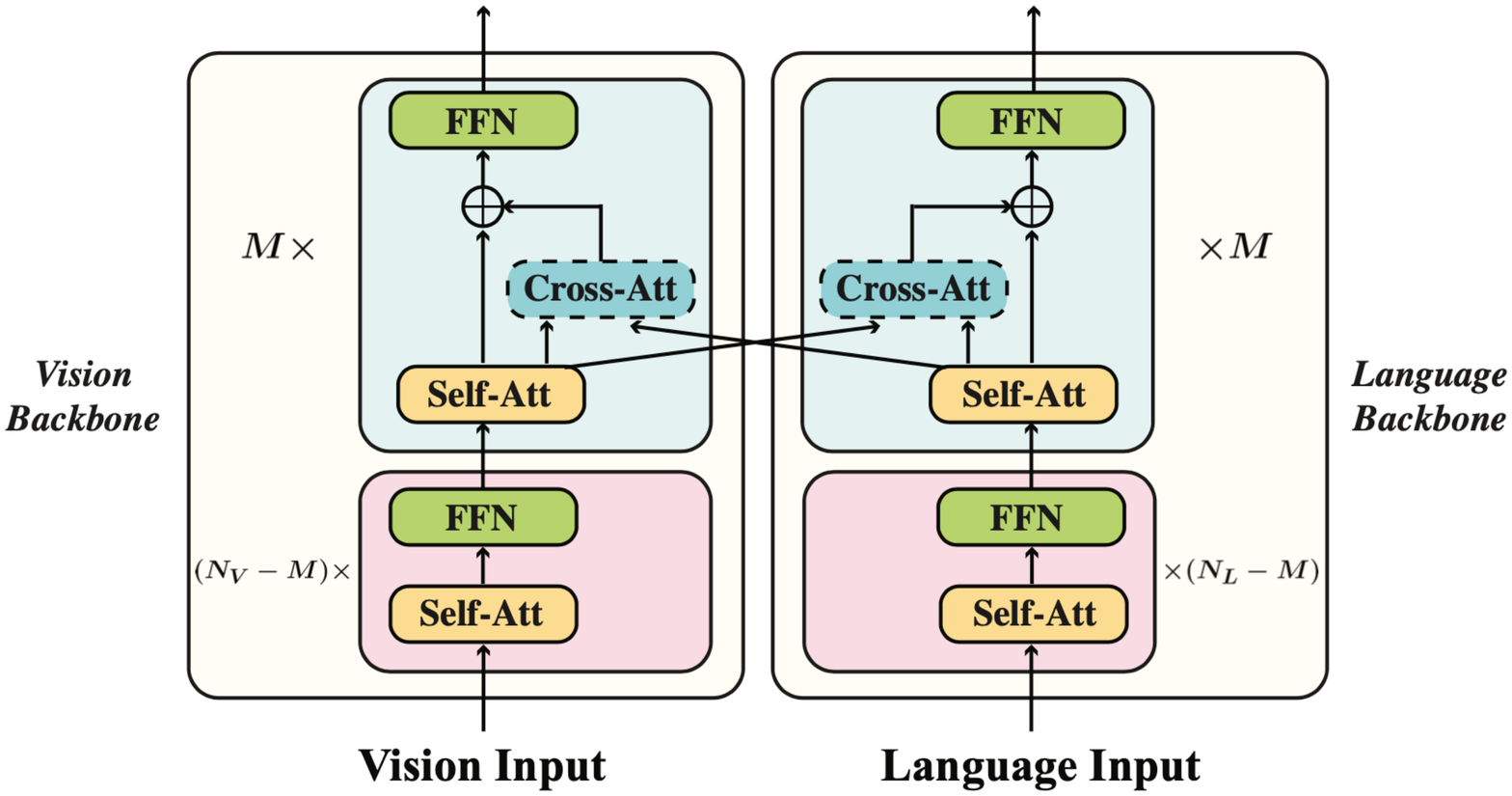}
\caption{\label{fig:FIBER_main_fig} Model architecture for \ModelName. Swin transformer is used as the image backbone, simplified here for illustration purposes.}
\vspace{-4.5mm}
\end{wrapfigure}

We answer this question by proposing two ideas: ($i$) a novel model architecture that can handle various types of tasks and pre-training strategies (high and low resolution inputs, image and region level outputs) more efficiently than previous work (see Section~\ref{sec:fiber} and~\ref{sec:exp}),
and ($ii$) a two-stage pre-training pipeline. 

In terms of \emph{architecture}, we present \ModelName, shown in Figure~\ref{fig:FIBER_main_fig}, which performs deep multimodal fusion in the backbone. Specifically, instead of having a few dedicated transformer layers on top of the image and text encoders for fusion (\emph{e.g.}, as is commonly done in previous work~\cite{li2020oscar, chen2020uniter, dou2021empirical, kamath2021mdetr, li2021grounded}), we propose to directly insert cross-attention modules into the image and text backbones. 
Additionally, we support the ability to switch between a dual encoder (for fast image retrieval) and a fusion encoder (for VQA and captioning) readily, by switching on or off the cross-attention modules. With the same model architecture, by simply adding an object detection head (\emph{e.g.}, Dynamic Head~\cite{dai2021dynamic}) on top, \ModelName can be readily extended to visual grounding, referring expression comprehension and (open-vocabulary) OD tasks as well.

\begin{figure*}
  \centering
\begin{subfigure}{0.24\linewidth}
    \includegraphics[width=1.0\linewidth]{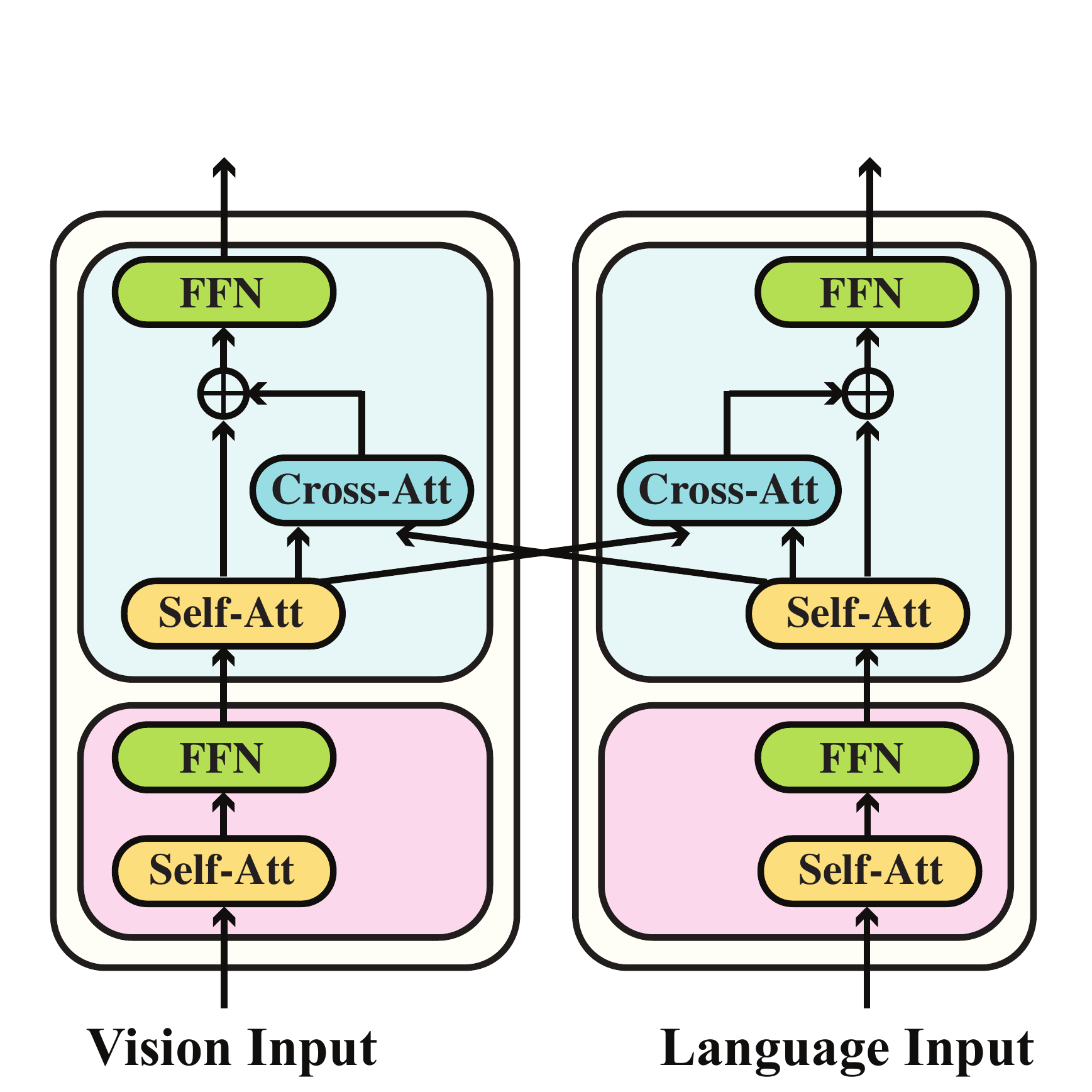}
    \caption{VQA/Visual Reasoning}
    \label{fig:taska}
  \end{subfigure}
   \begin{subfigure}{0.24\linewidth}
    \includegraphics[width=1.0\linewidth]{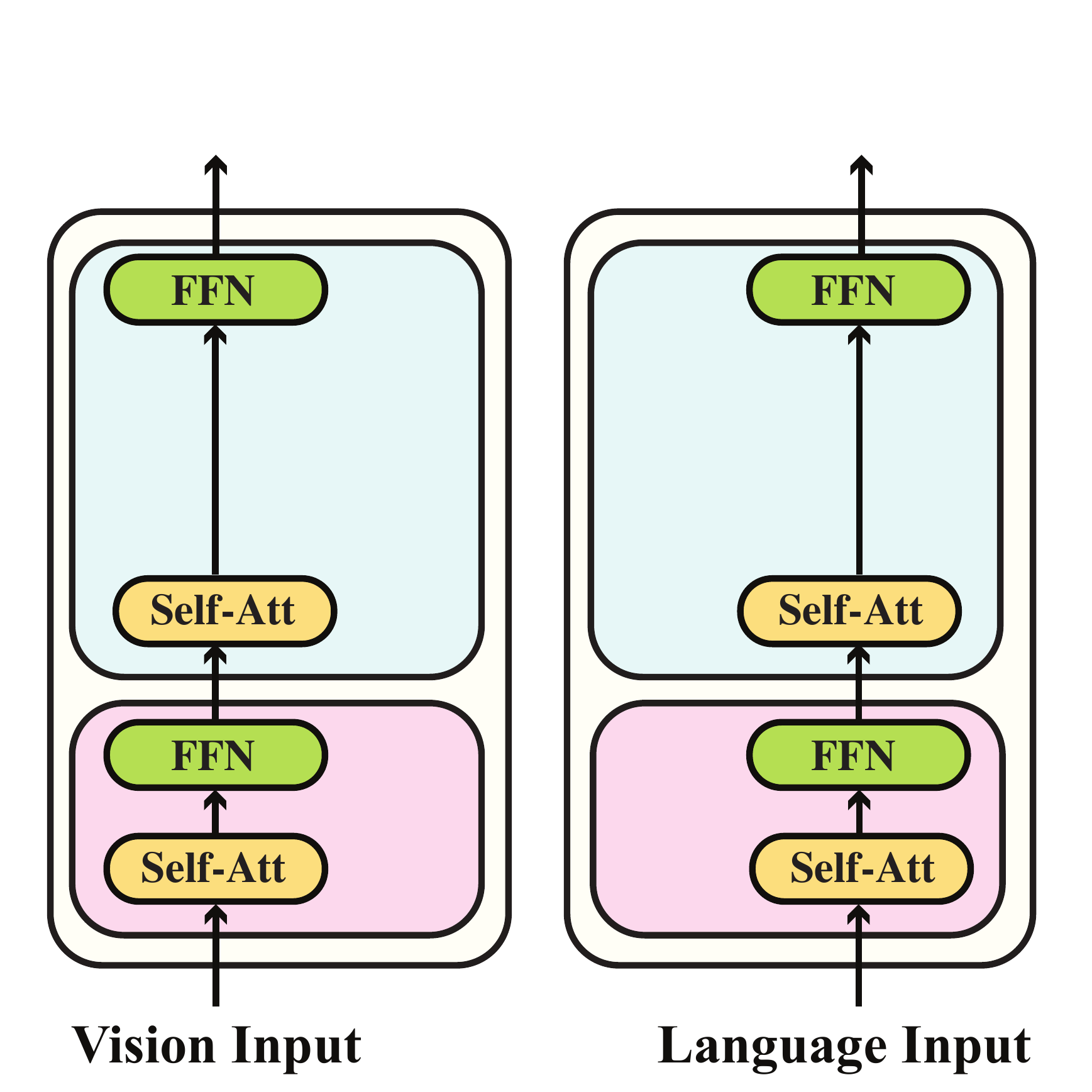}
    \caption{Retrieval}
    \label{fig:taskb}
  \end{subfigure}
  \begin{subfigure}{0.24\linewidth}
    \includegraphics[width=1.0\linewidth]{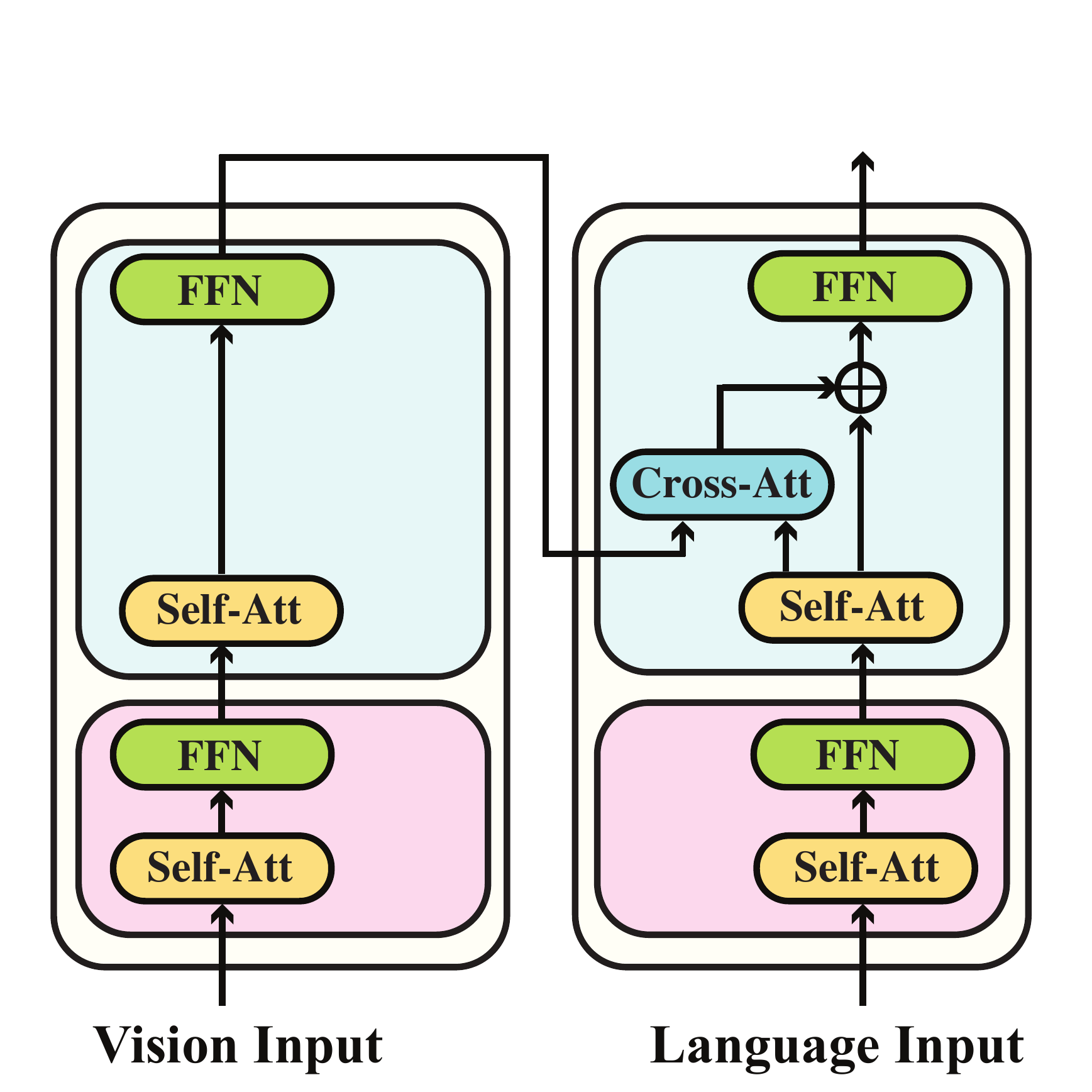}
    \caption{Captioning}
    \label{fig:taskc}
  \end{subfigure}
  \begin{subfigure}{0.24\linewidth}
    \includegraphics[width=1.0\linewidth]{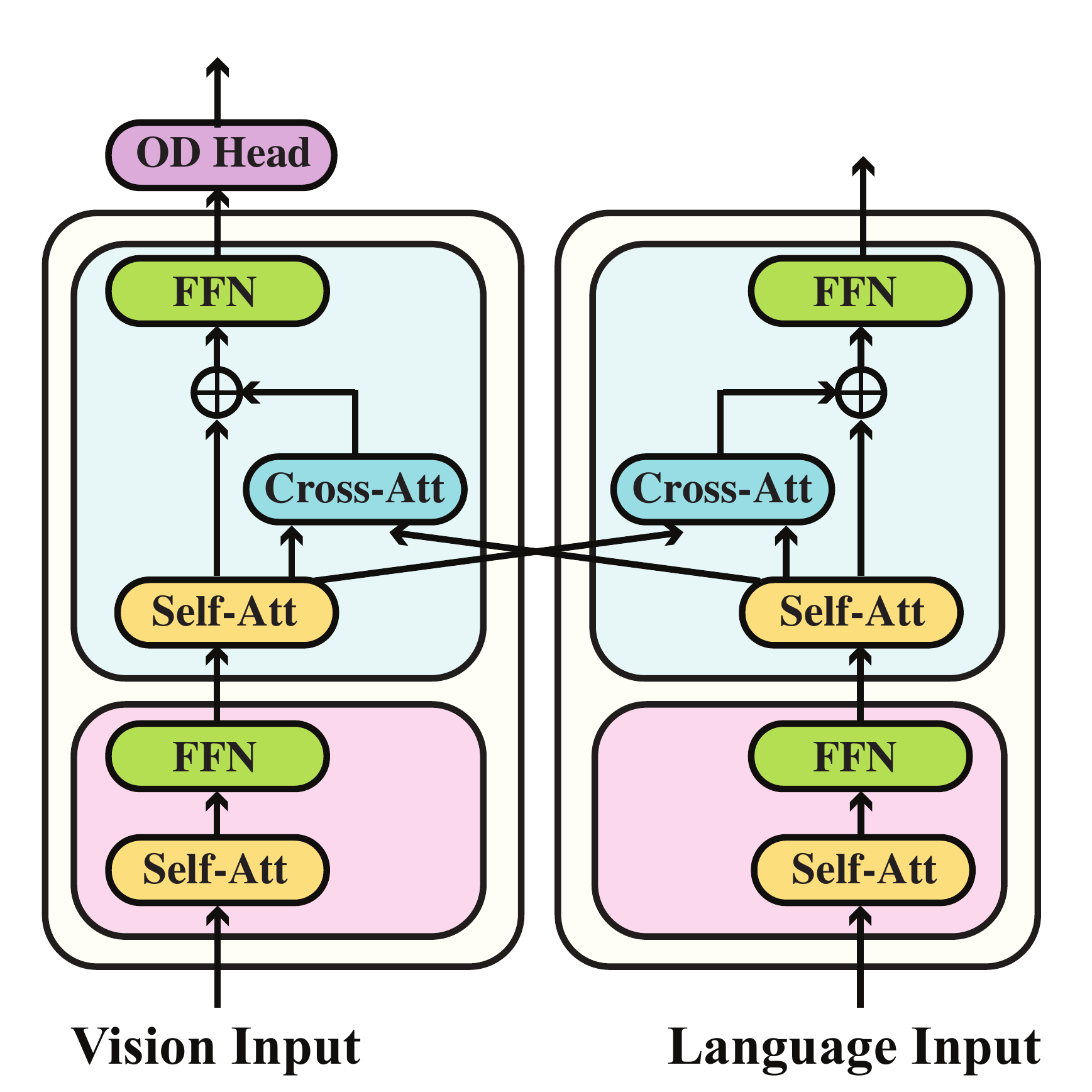}
    \caption{Grounding/OD}
    \label{fig:taskd}
  \end{subfigure}
  \caption{\ModelName can be readily adapted to various downstream VL tasks, ranging from VQA, image captioning and retrieval, to phrase grounding and object detection (OD). 
  \label{fig:FIBER_for_tasks}}
  \vspace{-2mm}
\end{figure*}

By considering the nature of different VL tasks, \ModelName is pre-trained with a coarse-to-fine two-stage pipeline, as detailed in Figure~\ref{fig:main}. Specifically,
\begin{itemize}[leftmargin=*]
    \item During \emph{coarse}-grained pre-training, \ModelName takes low-resolution (384$\times$384) images as input, and is pre-trained with image-text matching, masked language modeling, and image-text contrastive losses, as used in previous work~\cite{dou2021empirical,wang2021vlmo,wang2021ufo}. The pre-trained model can then be directly finetuned for VQA and image captioning tasks (Figure~\ref{fig:taska} and~\ref{fig:taskc}).
    By switching off the cross-attention modules, \ModelName also automatically functions as a dual encoder for fast image-text retrieval (Figure~\ref{fig:taskb}).
    \item During \emph{fine}-grained pre-training, \ModelName uses the coarse pre-trained model as initialization, in addition to randomly initialized parameters for the OD head. At this stage, the model takes high-resolution (800$\times$1,333) images as input, and is pre-trained with bounding box localization loss and word-region alignment loss, as used in GLIP~\cite{li2021grounded}. We use image-text-box data with ground-truth box annotations for pre-training, and the model can be directly fine-tuned for grounding and detection tasks (Figure~\ref{fig:taskd}).  
\end{itemize}
Compared to fine-grained pre-training, coarse-grained pre-training is easier to scale up, as it only requires paired image-text data which can be easily harvested from the web.
Crucially, we show that re-using all the parameters from our coarse-grained pre-trained model for fine-grained pre-training alleviates the requirement for large amounts of box-level annotated data. In our experiments, we show that on fine-grained tasks such as Flickr30k Entities, \ModelName using coarse-grained pre-training achieves gains even over previous SoTA (GLIP \cite{li2021grounded}) that uses 25$\times$ more box-level annotated images during the fine-grained pre-training stage. We also show that our architecture is much more efficient in terms of training time on OD tasks, as compared to GLIP .

\ModelName is the first end-to-end VLP model that can support VL tasks encompassing image-level and region-level outputs. We conduct experiments on VQAv2~\cite{antol2015vqa}, NLVR$^2$~\cite{suhr2018corpus}, COCO captioning~\cite{Lin2014MicrosoftCC}, NoCaps~\cite{agrawal2019nocaps}, COCO and Flickr30k image-text retrieval~\cite{plummer2015flickr30k}, as well as on phrase grounding~\cite{plummer2015flickr30k}, referring expression comprehension~\cite{yu2018mattnet}, COCO and LVIS detection~\cite{gupta2019lvis}, and a suite of 13 object detection in the wild datasets~\cite{li2021grounded}. We show that our model can provide consistent performance improvement over strong baselines (\emph{e.g.}, METER~\cite{dou2021empirical} and GLIP~\cite{li2021grounded}) across tasks.

%% file: subtex/2_related_work.tex
\section{Related Work}
\noindent\textbf{VLP for Classical VL Tasks.}
ViLBERT~\cite{lu2019vilbert} and LXMERT~\cite{tan-bansal-2019-lxmert} were the first two methods to introduce using transformers for VLP. Since then, we have witnessed a boom of VLP methods~\cite{li2019visualbert,li2019unicoder,su2019vl,yu2020ernie,hu2020vivo,yang2020tap,zhou2020unified,li2021scheduled,cho2021unifying,li2020unimo}. Early methods mainly focus on the use of pre-trained object detectors to extract image region features offline, such as UNITER~\cite{chen2020uniter}, OSCAR~\cite{li2020oscar}, VILLA~\cite{gan2020large} and VinVL~\cite{zhang2021vinvl}. More recently, end-to-end VLP methods that use the image directly as input have become popular. In these approaches, convolution networks or vision transformers~\cite{dosovitskiy2020image} are used as the image backbone, with additional transformer layers for modeling multimodal fusion~\cite{huang2020pixel,huang2021seeing,kim2021vilt,xue2021probing,li2021align,wang2021ufo}. Prominent examples along this line include ViLT~\cite{kim2021vilt}, ALBEF~\cite{li2021align}, SimVLM~\cite{wang2021simvlm}, METER~\cite{dou2021empirical}, X-VLM~\cite{zeng2021multi} and BLIP~\cite{li2022blip}. These models have achieved the current SoTA on major VL benchmarks such as VQA and image captioning. However, they cannot be directly used for tasks such as object detection. 

\vspace{0.3em}
\noindent\textbf{VLP for Vision Tasks.}
Recently, it has been shown that image-text data can be used to learn image encoders from scratch~\cite{desai2021virtex,sariyildiz2020learning}. By performing large-scale contrastive pre-training, CLIP~\cite{radford2021learning} and ALIGN~\cite{jia2021scaling} display strong zero-shot image classification capabilities. While these models mainly tackle image-level understanding tasks, MDETR \cite{kamath2021mdetr} extends the end-to-end OD model DETR \cite{carion2020end}, and uses contrastive learning along with an alignment loss to learn correspondences between image regions and text phrases, opening up the possibility to tackle tasks such as phrase grounding and long-tailed OD using VL models. This has inspired many follow-up works to further enhance the pre-training~\cite{li2021supervision,yao2021filip,mu2021slip,yang2022unified}, among which GLIP~\cite{li2021grounded} shows that OD can also be cast as a VL task (\emph{i.e.}, phrase grounding). However, it has not been shown how traditional VL tasks such as VQA, captioning and retrieval can be well supported in GLIP~\cite{li2021grounded} and MDETR~\cite{kamath2021mdetr}.

\vspace{0.3em}
\noindent\textbf{Unified VL Modeling.} There have been a few recent attempts that try to develop unified VL models. VL-T5~\cite{cho2021unifying} unifies VL tasks as text generation; however, pre-trained object detectors are used for image feature extraction, so the model cannot be end-to-end pre-trained.
UniT~\cite{hu2021unit} proposes a multimodal multi-task framework with a unified transformer; however, it can only support VQA and object detection tasks, but not captioning and grounding.
GPV~\cite{gupta2021towards} proposes a general-purpose vision system, and FLAVA~\cite{singh2021flava} presents a VL system similar to METER~\cite{dou2021empirical}; however, they did not evaluate on grounding and detection tasks, and their performance on other VL tasks is still far from SoTA. 
UniTAB~\cite{yang2021crossing} and OFA~\cite{wang2022unifying} reformulate grounding as a sequence generation task, by borrowing ideas from Pix2Seq~\cite{chen2021pix2seq}. However, these approaches have not been demonstrated to work on standard OD benchmarks, and also cannot be used as dual encoders for fast image retrieval. Our model is the first work that can support not only VQA, image captioning and $O(n+m)$ retrieval, but also visual grounding and object detection, with impressive performance across all tasks. A detailed comparison is provided in Table~\ref{tab:related}.
\input{tables/related_work}

%% file: tables/related_work.tex
\begin{table*}[t]
\setlength{\tabcolsep}{4pt}
\begin{center}
\small
\resizebox{1.0\textwidth}{!}
{
 \begin{tabular}{ccccccc} 
 \toprule
\textbf{Model} & \bf VQA$^\dagger$ & \bf $O(n+m)$ Retrieval$^\ddagger$ & \bf Captioning & \bf Grounding & \bf OD & \bf End2End \\
 \midrule
 ViLBERT~\cite{lu2019vilbert}, LXMERT~\cite{tan-bansal-2019-lxmert}, UNITER~\cite{chen2020uniter} & \checkmark & \ding{53} &  \ding{53} & \checkmark &   \ding{53} &   \ding{53}\\ 
 OSCAR~\cite{li2020oscar}, VinVL~\cite{zhang2021vinvl} & \checkmark & \ding{53} &  \checkmark & \checkmark &   \ding{53} &   \ding{53} \\
 PixelBERT~\cite{huang2020pixel}, CLIP-ViL~\cite{shen2021much}, ViLT~\cite{kim2021vilt} & \checkmark & \ding{53} & \ding{53} & \ding{53}  &\ding{53}& \checkmark\\ 
 CLIP~\cite{radford2021learning}$^*$, ALIGN~\cite{jia2021scaling} & \ding{53} & \checkmark &  \ding{53} & \ding{53} & \ding{53} & \checkmark \\
 VL-T5~\cite{cho2021unifying} & \checkmark & \ding{53} & \checkmark & \ding{53}  &\ding{53}& \ding{53} \\
 METER~\cite{dou2021empirical}, SimVLM~\cite{wang2021simvlm} & \checkmark & \ding{53} & \checkmark & \ding{53}  &\ding{53}& \checkmark\\
 ALBEF~\cite{li2021align}, FLAVA~\cite{singh2021flava}, VLMo~\cite{wang2021vlmo} & \checkmark  & \checkmark  & \ding{53} & \ding{53} & \ding{53} & \checkmark  \\
 BLIP~\cite{li2022blip}, CoCa~\cite{yu2022coca}, Flamingo~\cite{alayrac2022flamingo} &  \checkmark  & \checkmark  & \checkmark & \ding{53} & \ding{53} & \checkmark  \\
 MDETR~\cite{kamath2021mdetr}, GLIP~\cite{li2021grounded} &   \checkmark  & \ding{53}  & \ding{53} &  \checkmark &  \checkmark & \checkmark \\
 UniTAB~\cite{yang2021crossing},  X-VLM~\cite{zeng2021multi}, OFA~\cite{wang2022unifying} &   \checkmark  & \ding{53}  & \checkmark &  \checkmark &  \ding{53} & \checkmark \\
 \midrule
\ModelName & \checkmark & \checkmark & \checkmark & \checkmark & \checkmark & \checkmark \\
 \midrule
\end{tabular}
}
\caption{\textbf{Comparison among different VLP models}. \ModelName is the only VLP model that can support all tasks considered. ($\dagger$) VQA is used as a representative VL classification task. ($\ddagger$) $O(n+m)$ retrieval denotes model backbones process inputs $O(n+m)$ times given $n$ images and $m$ text sentences during image-text retrieval. ($*$) Here, we mainly focus on what tasks CLIP can be directly used for. 
}
\label{tab:related}
\vspace{-3mm}
\end{center}
\end{table*}

%% file: subtex/3_method.tex
\vspace{-3mm}
\section{Method}
\vspace{-2mm}
In this section, we first describe the proposed model architecture 
in Section~\ref{sec:fiber}. We then illustrate our two-stage pre-training paradigm in Section~\ref{sec:coarse-to-fine}, followed by fine-tuning strategies for all the tasks supported by \ModelName in Section \ref{downstream}.

\begin{wrapfigure}{R}{0.4\textwidth}
\centering
\vspace{-12mm}
\includegraphics[trim=5 40 20 20,clip,width=0.3\textwidth]{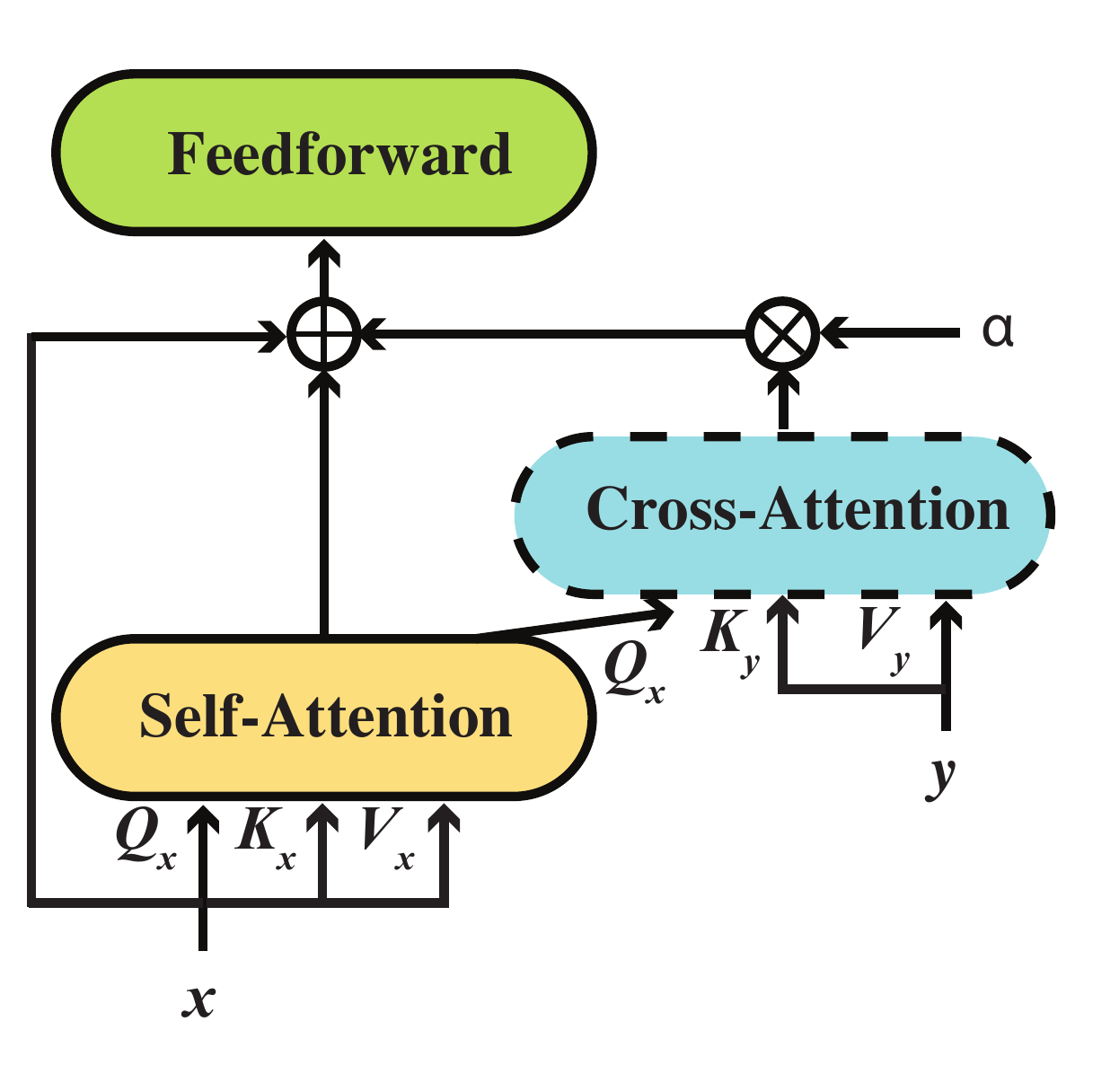}
\vspace{-3mm}
\caption{\label{fig:arch-ours} Illustration of performing fusion in the backbone. ($\xv$, $\yv$) are the (image, text) or (text, image) representations, and $\alpha$ is a learnable scalar.}
\vspace{-12mm}
\end{wrapfigure}

\subsection{Fusion in the Backbone}\label{sec:fiber}

The architecture of \ModelName is shown in Figure~\ref{fig:FIBER_main_fig}. Different from models that stack a modality fusion module on top of the vision or language backbones~\cite{chen2020uniter,dou2021empirical}, we insert multimodal fusion inside the backbones, and include a gating mechanism for the cross-modal layers (shown in Figure~\ref{fig:arch-ours}). 
Specifically, at each encoding layer, we have:
\begin{equation}
\begin{aligned}
        \tilde{\xv} &= \textsc{Self-Att}(\xv), \\
    \xv &= \xv + \tilde{\xv} + \alpha *  \textsc{Cross-Att}(\tilde{\xv}, \yv), \\
    \xv &= \xv + \textsc{FFN}(\xv),
\end{aligned}
\end{equation}
where $\alpha$ is a learnable parameter initialized to 0. For simplicity, we insert the same number of cross-attention layers into the vision and language backbones.

By inserting cross-attention layers with the gating mechanism, we enable cross-modal interactions without affecting the original computational flow of the backbones
at the beginning of model training.
Also, we can easily switch off the interactions by setting $\alpha$ to 0, and the backbones can be used in the dual-encoder setting. In addition, compared to stacking a large number of transformer layers on top of the backbones, our approach of inserting cross-attention layers is relatively light-weight and thus more memory-efficient.
To illustrate, both GLIP \cite{li2021grounded} and METER~\cite{dou2021empirical} use an additional 110M modality fusion parameters for a base-size model, while \ModelName only adds about 26M parameters. During training, the fusion module of \ModelName only consumes half of the FLOPs needed by METER (12.35 vs. 24.04 GFLOPs for one instance).
We experimented with two other model variants for fusion in the backbone, the details of which are provided in Appendix.

\subsection{Coarse-to-Fine Pre-training}\label{sec:coarse-to-fine}
We divide VL tasks into two categories based on whether or not we need to generate region-level outputs on the image side. 
While these two kinds of tasks are characteristically different, they both require fusion between the vision and language modalities, and we hypothesize that sharing as many parameters as possible between the model used for these two sets of tasks will be beneficial. Based on this motivation, we propose a two-stage pre-training paradigm, where we first pre-train models with image-level objectives on images at low resolution, and then perform further pre-training with region-level objectives where the input images are at a higher resolution. In this way, the coarse-grained supervision from the first stage can provide good initialization for the second stage for all the shared parameters. \ModelName with the same architecture (Swin Transformer \cite{liu2021swin} and RoBERTa \cite{liu2019roberta}) is used as the backbone for both stages of pre-training. 

\paragraph{Coarse-grained Pre-training.} For tasks like VQA and captioning, it has been demonstrated~\cite{li2021align,dou2021empirical,wang2021vlmo} that masked language modeling (MLM), image-text matching (ITM), and image-text contrastive (ITC) objectives are helpful for ViT-based VLP models. Following previous work, we use all the three objectives during pre-training. Specifically, 
\begin{itemize}[leftmargin=*]
    \item \textbf{For ITC}, the inserted
    cross-attention modules are switched off, so \ModelName functions as a dual encoder. Given a batch of $N$ image-caption pairs, we first compute their representations with our vision and language encoders independently without modality fusion, and then maximize the similarities between $N$ positive image-text pairs while minimizing the similarities between the rest $N^2-N$ negative pairs, via a contrastive loss. 
    \item \textbf{For MLM and ITM}, the inserted cross-attention modules are switched on, so \ModelName now functions as a fusion encoder. For MLM, we randomly mask 15\% of the input tokens and the model is trained to reconstruct the original tokens. For image-text matching, the model is given an image-text pair and predicts whether they are matched. Following VLMo~\cite{wang2021vlmo}, we sample global hard negatives based on the similarities computed from the above ITC loss.
\end{itemize}

\paragraph{Fine-grained Pre-training.}
Most existing VL architectures \cite{chen2020uniter, tan-bansal-2019-lxmert, kamath2021mdetr, li2022blip,wang2022unifying, cho2021unifying} use vanilla transformers both for encoding the vision as well as language inputs. However, in contrast to tokens in text, the entities of interest in images do not all occur at the same scale. Being able to accurately model the image at different scales is especially important for tasks such as object detection and phrase grounding. To handle this, it is typical in object detection literature to use input images at higher resolutions  
(800$\times$1333), which becomes problematic when using vanilla transformers that scale quadratically in the length of the input sequence. As mentioned earlier, we use a Swin Transformer \cite{liu2021swin} as our image encoder, which provides hierarchical representations of the image while having linear complexity in the size of the image. We combine these multi-scale representations using an FPN \cite{Lin2017FeaturePN} for object detection training. For fine-grained pre-training, we switch on the cross-attention modules, using \ModelName as a fusion encoder. This ensures that the image representations that are passed to the FPN are already text-aware, and is a crucial difference compared to GLIP~\cite{li2021grounded}, where the image-text fusion takes place in the object detection head.  Once the text-aware image features are extracted by the Swin backbone and image-aware text features are extracted using RoBERTa \cite{liu2019roberta}, the image features after the FPN are fed to a DynamicHead~\cite{dai2021dynamic}
which predicts a set of regions. Just as in \cite{li2021grounded}, we compute the dot product between the image region features $\rv$  and the contextualized token representations $\tv$ to compute the grounding score:
\begin{equation}
\begin{aligned}
        \iv, \tv = \ModelName(I, T),\,\,
        \rv = \textsc{OD-Head}(\iv), \,\,
    S_{\textsc{grounding}} = \rv \tv^\top,
\end{aligned}
\end{equation}
where $\rv$ represents regions that are text aware, produced using the OD-Head that takes as input $\iv$, which are image representations that are already text-aware
and 
$\tv$ are the text features that have already attended to the image features.
The typical object detection model has a classification head that predicts the label of the object, and a localization head that predicts the bounding box. We follow GLIP~\cite{li2021grounded} by substituting the classification head with the grounding score $S_{\textsc{grounding}}$. The localization loss is composed of two parts: a centerness loss and GIoU loss, which are used to supervise the box prediction. Taken together, \ModelName learns the correspondence between regions in the image and phrases in the text, making it possible to tackle tasks such as phrase grounding and object detection using the same framework.  We use ATSS framework \cite{Zhang2020BridgingTG} in our paper, but our method can be combined easily with other object detectors such as Faster-RCNN \cite{ren2015faster} and RetinaNet \cite{Lin2020FocalLF} as well.

\subsection{Adaptation to Downstream Tasks}\label{downstream}
 We now describe how we adapt \ModelName to different downstream tasks as depicted in Figure~\ref{fig:FIBER_for_tasks}.
\begin{itemize}[leftmargin=*]
    \item \textbf{For VL classification tasks such as VQA}, we use \ModelName as a fusion encoder. Specifically, the top $M$ layers of the vision and language backbones interact with each other and produce multimodal representations. The final layer representations of the two modalities are concatenated together to generate the final outputs for tasks such as VQA and visual reasoning.
    \item \textbf{For retrieval tasks}, 
    we switch off the inserted cross-attention modules to use \ModelName as a dual encoder for fast image-text retrieval. 
    \item \textbf{For captioning},
    we adapt \ModelName 
    by only keeping the
    image-to-text cross-attentions and using causal masks in the decoding side. The representations of the final image encoding layer are fed into the cross-attention modules. In this way, the model is turned into a seq2seq  model~\cite{sutskever2014sequence,cho2014learning} and performs captioning in an auto-regressive way.
\item \textbf{For phrase grounding, object detection and referring expression comprehension}, we use \ModelName as a fusion encoder, and the OD-Head introduced during fine-grained pre-training receives image features that are already language aware due to the multimodal representations extracted by \ModelName. The pre-trained model is directly used without any modifications for these tasks.

\end{itemize}

%% file: subtex/4_experiments.tex
\section{Experiments}
\label{sec:exp}
\input{tables/od_efficiency}
\paragraph{Pre-training Datasets.} Following previous work~\cite{chen2020uniter,kim2021vilt,li2021align,dou2021empirical,wang2021ufo,wang2021vlmo}, we perform coarse-grained pre-training on COCO~\cite{Lin2014MicrosoftCC}, Conceptual Captions~\cite{sharma2018conceptual}, SBU Captions~\cite{ordonez2011im2text},  and Visual Genome~\cite{krishna2017visual}. The four datasets consist of about 4M images in total. For fine-grained pre-training, we use two data sources: data curated by MDETR \cite{kamath2021mdetr} after removing the COCO images, and the Objects365 \cite{Shao2019Objects365AL} detection dataset, together consisting of about 0.8M images. We ensure that we exclude any data that exists in the validation or test splits of downstream tasks.

\noindent\textbf{Architecture.} We adopt Swin-Base~\cite{liu2021swin} and RoBERTa-Base~\cite{liu2019roberta} as our vision and text backbones, which are initialized with weights from uni-modal pre-training.
We insert cross-attention blocks into the top 6 blocks of the vision and text encoders. The input resolution is $384\times384$ for coarse-grained pre-training and $800\times 1,333$ for fine-grained pre-training.
Using a hierarchical vision transformer enables us to efficiently tackle these high resolution tasks, which would be expensive in models such as BLIP \cite{li2022blip} that rely on the vanilla transformer architecture. In METER \cite{dou2021empirical}, which does explore using a Swin transformer as the image encoder, the multi-modal fusion occurs in layers specifically designed to align the modalities, only after the image and text features are extracted from the uni-modal backbones. This is in contrast to our approach where the hierarchical image features that are used in the FPN for fine-grained training are already language aware, due to the multi-modal fusion being in the backbone. 
This also lets us avoid adding additional ``language-aware deep fusion layers'' \cite{li2021grounded} as part of the OD head as in GLIP, resulting in 1.5x faster training while maintaining performance as shown in Table \ref{tab:COCO timing analysis}.
While in principle it would be possible to use the image features extracted by METER's backbone for object detection, it would be necessary as in GLIP to add additional layers to make the visual features ``language-aware'' for good detection performance, especially on datasets with limited training data and with rare and infrequent objects.

\vspace{0.3em}
\noindent\textbf{Implementation Details.} We perform coarse-grained pre-training for 100k steps with 4,096 batch size on 64 A100 GPUs. We use AdamW~\cite{loshchilov2018decoupled} with the peak learning rates of 1e-4 for the backbones and 5e-4 for the cross-modal parameters. We use linear warmup over the first 1k steps and linear decay. For fine-grained pre-training, we train for 800k steps on 64 V100 GPUs, with a batch size of 64. We use a learning rate of 1e-5 for the language backbone, and 1e-4 for the rest of the model with a weight decay of 0.01. We use a linear warmup over the first 2k steps and then a constant learning rate, with two learning rate drops by a factor of 10 at 67\% and 89\% of the total number of steps.

\subsection{Results on Downstream Tasks}

\input{tables/results_vl}
\noindent\textbf{Vision-Language Classification.} We first experiment on two representative VL classification tasks, including VQAv2~\cite{antol2015vqa} and NLVR$^2$~\cite{suhr2018corpus}. 
As reported in Table~\ref{tab:results1}, we achieve the best performance compared to other models in the same setting. It is worth noting that \ModelName pre-trained with 4M images can achieve better performance than BLIP trained with 129M images and SimVLM trained with 1.8B images. The results indicate that introducing fusion modules into the backbone is an effective alternative to appending them on the top of uni-modal backbones.

\vspace{0.3em}
\noindent\textbf{Image-Text Retrieval.} 
In Table \ref{tab:results1} we report image retrieval performance in the dual encoder setting, achieving competitive performance  on both Flickr30k~\cite{plummer2015flickr30k} and COCO~\cite{Lin2014MicrosoftCC} retrieval tasks.
However, previous work has shown that fusion encoders obtain superior performance, albeit at the cost of efficiency as it involves feeding every image-text pair into the model.  To illustrate, on the COCO test data, ranking the similarities between 5K images and 25K captions requires the model to process each image-caption pair 75M times, whereas the dual encoder model only needs 30K forward passes. As shown in Table~\ref{tab:results_retrieval}, the fusion encoder can indeed surpass the dual encoder on retrieval tasks by a large margin. In addition, directly ensembling the two models by summing their similarity scores together for each image-caption pair can bring us huge improvements.

Further, we explore combining the strengths of both strategies by performing re-ranking as in~\cite{geigle2022retrieve, li2022blip, li2021align}. Specifically, we first retrieve the top-$k$ most similar instances using the dual encoder setup, and then add the similarity scores between the given instance and the top-$k$ candidates provided by the fusion encoder to the original scores to perform retrieval. From Table~\ref{tab:results_retrieval}, we can see that this strategy provides a balance between efficiency and performance, and that just re-ranking the top-10 instances can achieve comparable performance with ensembling.
\input{supp_tables/retrieval}


\input{tables/results_captioning}

\vspace{0.3em}
\noindent\textbf{Image Captioning.} We also evaluate our models on COCO~\cite{Lin2014MicrosoftCC} and NoCaps~\cite{agrawal2019nocaps} captioning to test whether \ModelName can be adapted to generation tasks. As in Table~\ref{tab:caption}, \ModelName can achieve better performance than models trained on the same data with and without CIDEr optimization~\cite{rennie2017self}. We find that integrating GOLD~\cite{pang2021text} into \ModelName can bring significant improvements, outperforming models trained with hundreds of millions of images.
Notably, we establish the absolute state-of-the-art CIDEr scores on COCO for base-size models. Considering that \ModelName is not pre-trained to perform captioning, the results demonstrate the strong generalization ability of \ModelName.

\noindent\textbf{Phrase Grounding.}
Our fine-grained pre-training stage incorporates Flickr30k entities grounding data, and we achieve 87.4 on the Recall@1 metric on the test set without any subsequent fine-tuning. This not only surpasses the current SoTA \cite{li2021grounded} using a smaller sized model (Swin-B compared to their Swin-L), but also uses 25x less fine-grained data. Our \ModelName model is able to leverage the image-text coarse-grained pre-training stage better, instead of relying on expensive pseudo-labelling of large web-scale corpus and subsequent high-resolution training on this generated fine-grained data as in \cite{li2021grounded}. We also compare our approach without using any coarse-grained VL training (image encoder initialized to Swin-B weights from IN22k, and text encoder initialized to pre-trained RoBERTa), and even in this setting, we are able to outperform a similarly sized GLIP model (GLIP-B), proving that our fusion in the backbone is better at capturing fine-grained image-text understanding.

\input{tables/results_phrase_grounding}

\vspace{0.3em}
\begin{wrapfigure}{R}{0.4\textwidth}
\centering
\vspace{-6mm}
\includegraphics[width=0.38\textwidth]{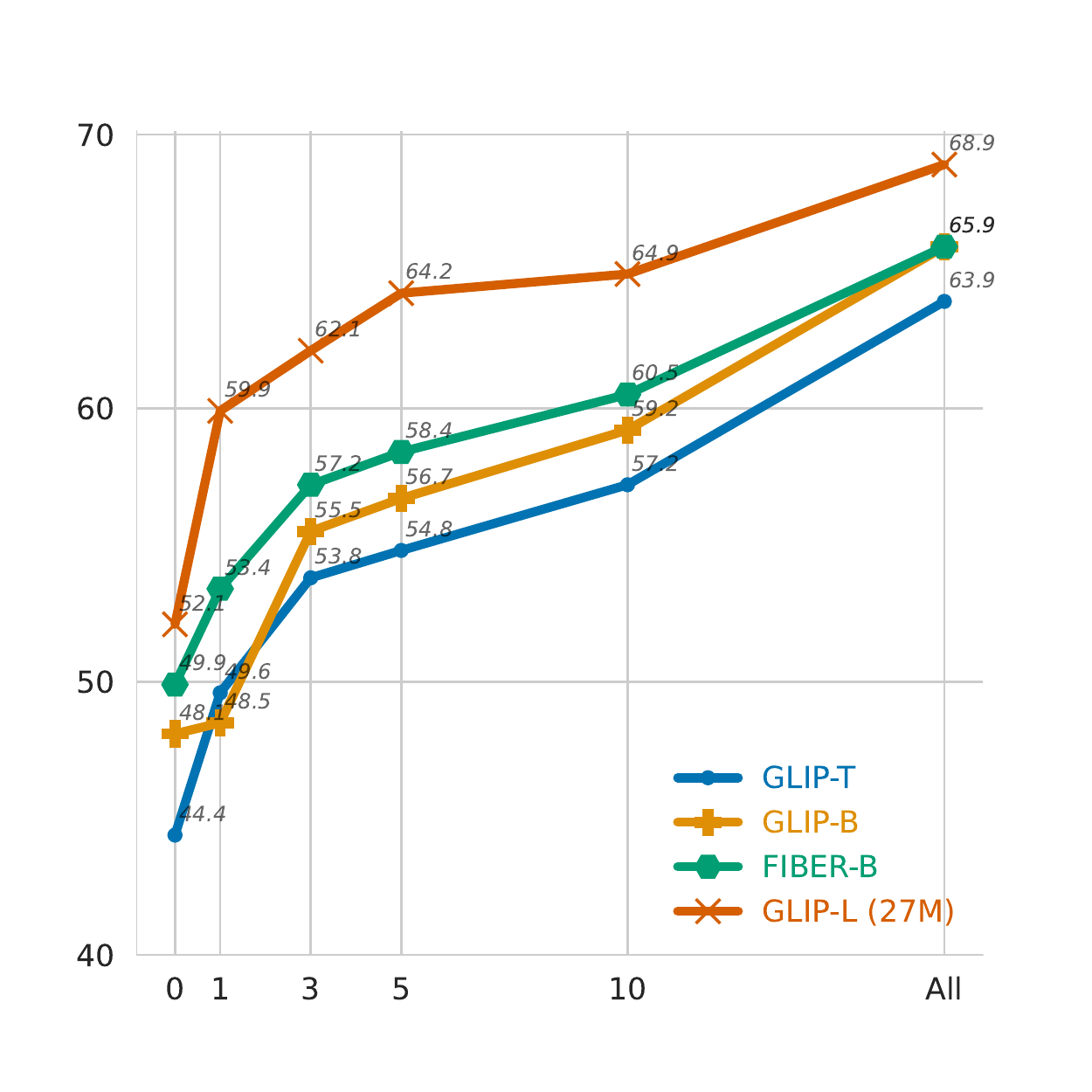}
\vspace{-3mm}
\caption{\label{fig:odinw-fewshot} Few-shot results on the aggregated 13 ODinW datasets.} 
\end{wrapfigure}

\noindent\textbf{Referring Expression Comprehension (REC).}
In contrast to many previous works \cite{chen2020uniter, gan2020large, lu2019vilbert} that tackle the REC task by re-ranking object proposals provided by an off-the-shelf detector, we follow \cite{kamath2021mdetr} to directly predict the bounding box for the given referring expression.
Using our proposed two stage pre-training, \ModelName achieves better performance than current SoTA \cite{wang2022unifying} that uses a Large sized model. Notably, on RefCOCOg \cite{Yu2016ModelingCI}, which contains much longer referring expressions than in RefCOCO/RefCOCO+ \cite{kazemzadeh2014referitgame}, we observe more than 2 points boost over OFA-L. On the challenging testB split of both RefCOCO and RefCOCO+, \ModelName outperforms current SoTA, OFA-L.

\input{tables/results_referring_expression}
\input{tables/results_object_detection}

\paragraph{Object Detection.}
We report \ModelName results on two standard object detection benchmarks, COCO \cite{Lin2014MicrosoftCC} and LVIS \cite{gupta2019lvis}, in zero-shot transfer\footnote{Following \cite{radford2021learning, Zhai2021LiTZT}, we consider zero-shot transfer to mean that during pre-training we may have seen relevant data but it is not used for training for the task of interest. For instance, our coarse-grained pre-training includes some images from COCO (without any box information), but we do not have any COCO images in our fine-grained training that we use to train the object detection head.} as well as fine-tuned settings in Table \ref{tab:ZS/finetuned_OD}.
The LVIS dataset consists of a long-tail of object classes, and is a popular test-bed for evaluating models on their generalization capabilities and robustness to class imbalance. On the APr metric, which is the Average Precision on rare objects, \ModelName outperforms GLIP-L which is a bigger model and also trained with 25$\times$ more fine-grained data. 

We also report zero-shot and fine-tuned results on a suite of 13 ODinW (object detection in the wild) datasets, spanning various domains and show consistent performance improvements over previous SoTA. Additionally, in Figure \ref{fig:odinw-fewshot}, we report few-shot results aggregated across these 13 datasets and show better data efficiency over GLIP-B trained with the same fine-grained data.

\paragraph{Ablation Study.} In Appendix~\ref{sec:ablation} and \ref{sec:additional_results}, we have provided detailed ablations that guided our architecture design, including ablations on fusion strategies, pre-training objectives, architecture for captioning, and additional results on open-ended VQA, and detailed few-shot ODinW results. Due to the space limit, these ablations and additional results are only provided in the Appendix. Some important observations are summarized below. 
($i$) Co-attention works similarly to merged attention for fusion in the backbone.
($ii$) Adding a gating parameter in co-attention allows the addition of fusion in more layers, and also gives better performance than merged attention.
($iii$) Adding co-attention in the last 6 layers provides a balance between performance and efficiency.
($iv$) MLM, ITM with hard negative mining, and ITC are all important pre-training objectives for training FIBER-style models.

%% file: tables/od_efficiency.tex
\begin{wraptable}{R}{0.5\textwidth}
\vspace{-5mm}
\setlength{\tabcolsep}{4pt}
\begin{center}
\small
\begin{tabular}{cccc}
 \toprule
 \bf Type of &  \bf COCO     & \bf GPU-hours  &  \multirow{2}{*}{\bf Sec/Iter} \\
 \bf Fusion              &  \bf Val2017  &  \bf V100 (32GB)              &          \\
 \midrule
No Fuse & 53.9 & 511 & 1.31 \\
GLIP-B \cite{li2021grounded} & 54.6 & 840 & 2.14 \\
\midrule
\ModelName-B & 54.5  & 540 &  1.38\\
 \midrule
\end{tabular}
\caption{Object detection on COCO \cite{Lin2014MicrosoftCC}, without vision-language pre-training. We initialize the text encoder and image backbones using a pre-trained RoBERTa and a Swin transformer pre-trained on ImageNet22k. Our proposed \ModelName model achieves the same performance as GLIP~\cite{li2021grounded} while taking much less time to train. 
\label{tab:COCO timing analysis}}
\end{center}
\vspace{-5mm}
\end{wraptable}

%% file: tables/results_vl.tex
\begin{table*}[t]
\setlength{\tabcolsep}{2.5pt}
\begin{center}
\small
{
 \begin{tabular}{ccccccccccccccccc} 
 \toprule
    \multirow{2}{*}{\bf Model}  &  \multirow{2}{*}{\textbf{\tabincell{c}{\#Pretrain \\ Images}}}  &  \multicolumn{2}{c}{\bf VQAv2} & \multicolumn{2}{c}{\bf NLVR$^2$} &  \multicolumn{2}{c}{\bf Flickr30k} &  \multicolumn{2}{c}{\bf COCO}  & \\ 
    \cmidrule(lr){3-4} \cmidrule(lr){5-6} \cmidrule(lr){7-8} \cmidrule(l){9-10}
     & & \bf test-dev & \bf test-std & \bf dev & \bf test-P & \bf IR@1 & \bf TR@1  & \bf IR@1 & \bf TR@1 \\ 
     \midrule
       \multicolumn{9}{l}{ { \it{Base-size models pre-trained on COCO, VG, SBU, and CC datasets} } }\\
       \midrule
    UNITER-B~\cite{chen2020uniter} & 4M &  72.70 & 72.91 & 77.18 & 77.85 & 72.5 & 85.9 & 50.3 & 64.4 & \\
    VILLA-B~\cite{gan2020large} & 4M & 73.59 & 73.67 & 78.39 & 79.30 & 74.7 & 86.6 & - & - &\\
    UNIMO-B~\cite{li2020unimo} & 4M & 73.79 & 74.02 & - & - & - & -  & - & - &\\
    ViLT-B~\cite{kim2021vilt} & 4M & 71.26 & - & 75.70 & 76.13 & 64.4 & 83.5 & 42.7 & 61.5 & \\
    ALBEF-B~\cite{li2021align} & 4M &  74.54 & 74.70 & 80.24 & 80.50 & \demph{82.8}$^\dagger$ &  \demph{94.3}$^\dagger$ & \demph{56.8}$^\dagger$ & \demph{73.1}$^\dagger$ & \\
  VLMo-B~\cite{wang2021vlmo} & 4M &  76.64 & 76.89& 82.77 & 83.34 & 79.3 & 92.3 & 57.2 & 74.8\\
  METER-Swin-B~\cite{dou2021empirical} & 4M &  76.43 & 76.42 & 82.23 & 83.47 & 79.02 & 92.4 & 54.85 & 72.96 \\
  X-VLM~\cite{zeng2021multi} & 4M &  78.22 & 78.37 & 84.41 & 84.76 & \demph{86.9}$^\dagger$ & \demph{97.0}$^\dagger$ & \demph{63.4}$^\dagger$ & \demph{81.2}$^\dagger$ \\
  \midrule
       \multicolumn{9}{l}{ \demph{ \it{Models pre-trained on more data and/or with larger size} } }\\
       \midrule
       \demph{VLMo-L~\cite{wang2021vlmo}}  & \demph{4M} & \demph{79.94} & \demph{79.98} & \demph{85.64} & \demph{86.86} & \demph{84.5} & \demph{95.3} & \demph{60.6} & \demph{78.2} \\
       \demph{BLIP$_{\text{CapFilt-L}}$~\cite{li2022blip}}  & \demph{129M} &  \demph{78.25} & \demph{78.32}& \demph{82.15} & \demph{82.24} & \demph{87.5$^\dagger$} & \demph{97.2$^\dagger$} & \demph{64.1$^\dagger$} & \demph{81.2$^\dagger$} & \\
       \demph{SimVLM-B~\cite{wang2021simvlm}} & \demph{1.8B} & \demph{77.87}& \demph{78.14}& \demph{81.72} & \demph{81.77} & \demph{-} & \demph{-} & \demph{-}& \demph{-}& \\ 
       \demph{SimVLM-H~\cite{wang2021simvlm}} & \demph{1.8B} &  \demph{80.03} & \demph{80.34} & \demph{84.53}& \demph{85.15}& \demph{-} & \demph{-} & \demph{-}& \demph{-} \\
  \midrule
  \ModelName-B & 4M & \textbf{ 78.55} & \textbf{78.46} & \textbf{84.59} & \textbf{85.52} & \bf 81.44 & \bf 92.90 &\textbf{58.01} & \textbf{75.38} \\ 
  \midrule
  \end{tabular}
  }
  \vspace{-2mm}
  \caption{Results on VL classification and retrieval. We also include models pre-trained on more data and/or with larger size. \ModelName and VLMo use dual encoders for retrieval. ($\dagger$) ALBEF, X-VLM, and BLIP first use its dual encoder to
obtain top-$k$ candidates, and then use its fusion encoder to re-rank the
candidates. Our retrieval results with re-ranking are provided in Table~\ref{tab:results_retrieval}. All the other models use fusion encoders.  }
  \label{tab:results1}
   \end{center}
   \vspace{-6mm}
\end{table*}

%% file: supp_tables/retrieval.tex
\begin{table*}[t]
\setlength{\tabcolsep}{2.0pt}
\begin{center}
\small
\resizebox{\textwidth}{!}
{
 \begin{tabular}{ccccccccccccccccccccccc} 
 \toprule
    \multirow{2}{*}{\bf Model}  &   \multicolumn{6}{c}{\bf Flickr30k} &  \multicolumn{6}{c}{\bf COCO}  & \\ 
    \cmidrule(lr){2-7} \cmidrule(lr){8-13} 
     &  \bf IR@1 & \bf IR@5 & \bf IR@10  & \bf TR@1 & \bf TR@5 & \bf TR@10 &  \bf IR@1 & \bf IR@5 & \bf IR@10  & \bf TR@1 & \bf TR@5 & \bf TR@10 \\ 
  \midrule
  \ModelName-ITC &  81.44 & 96.72 & 98.48 & 92.90 & 99.50 &  99.90 & 58.01 & 83.45 & 90.11 & 75.38 & 94.04 & 97.36 \\ 
  \ModelName-ITM & 84.10 & 97.54 & 98.88 & 95.10 & 99.60 & 99.90 & 59.03 & 84.04 & 91.03 & 75.14 & 93.88 & 97.36\\
  \ModelName-ITC+ITM Ensemble & 90.96 & 98.44 & 99.14 & 96.00 & 99.70 & 100.00 & 69.73 & 90.66 & 94.59 & 80.10 & 95.60 & 97.98\\
  \midrule
  ALBEF~\cite{li2021align} & 82.8 & 96.7 & 98.4 & 94.3 & 99.4 & 99.8 & 56.8 & 81.5 & 89.2 & 73.1 & 91.4 & 96.0 \\
 X-VLM~\cite{zeng2021multi} & 86.1 & 97.4 & 98.7 & \bf 96.8 & \bf 99.8 & 100.0 & 63.1 & 85.7 & 91.6 & \bf 80.4 & 95.5 & \bf 98.2 \\
  \ModelName-Rerank-10 & 90.94 & 98.16 & 98.48 & 95.80 & 99.60 & 99.90 & 68.71 & 87.69 & 90.09 & 79.66 & 95.34 & 97.36 \\
  \ModelName-Rerank-20 & 90.10 & 98.38 & 99.14 & 95.90 & \bf 99.80 & 100.00 & 69.32 & 89.52 & 93.33 & 79.78 & 95.20 & 97.66\\
  \ModelName-Rerank-50 & \bf 91.08 & 98.50 & \bf 99.37 & 96.10 & 99.70 & 100.00 & 69.58 & 90.41 & 94.35 & 79.98 & 95.40 & 97.76\\
  \ModelName-Rerank-100 & 91.02 & \bf 98.54 & 99.34 & 96.00 & 99.70 & 100.00 & \bf 69.63 & \bf 90.54 & \bf 94.47 & 80.06 & \bf 95.60 & 97.96\\
  \midrule
  \end{tabular}
  }
  \vspace{-2mm}
  \caption{Additional results on image-text retrieval, where ($i$) the fusion encoder is used for retrieval, or ($ii$) the dual encoder is first used to obtain top-$k$ candidates, and then the fusion encoder is used to re-rank the candidates. We also provide a full set of results on all evaluation metrics.}
  \label{tab:results_retrieval}
   \end{center}
   \vspace{-3mm}
\end{table*}

%% file: tables/results_captioning.tex
\begin{table*}[t]
\setlength{\tabcolsep}{6pt}
\begin{center}
\small
\def \arraystretch{0.95}
 \begin{tabular}{cccccccccccccc} 
 \toprule
    \multirow{2}{*}{\bf Model}  & \multirow{2}{*}{\textbf{\tabincell{c}{\#Pretrain \\ Images}}}  & \multicolumn{4}{c}{\bf COCO} & \multicolumn{2}{c}{\bf NoCaps Val} & \multicolumn{2}{c}{\bf NoCaps Test}  \\
    \cmidrule(lr){3-6} \cmidrule(lr){7-8} \cmidrule(l){9-10}
     & & \bf B@4 & \bf M & \bf C &\bf S & \bf C & \bf S & \bf C & \bf S  \\
    \midrule
     \multicolumn{9}{l}{  \it{Models trained without CIDEr optimization} } \\
     \midrule
    UFO-B~\cite{wang2021ufo} & 4M & 36.0 & 28.9 &  122.8 & 22.2 & 80.7 &12.5 & 78.8 & 12.5 \\
    ViTCAP~\cite{fang2022injecting} & 4M & 36.3 & 29.3 & 125.2 & 22.6 & - & - & - & -\\
    METER-CLIP-B~\cite{dou2021empirical} & 4M & 38.8 & 30.0 & 128.2 & 23.0 & - & - & - & - \\
    X-VLM~\cite{zeng2021multi} & 4M  &  39.8 & - & 133.1 & - & - & - & - & -\\
    VinVL-B~\cite{zhang2021vinvl} & 5.7M & 38.2 & 30.3 & 129.3 & \bf 23.6 & - & - & - & -\\
      \demph{BLIP$_{\text{CapFilt-L}}$~\cite{li2022blip}} & \demph{129M} & \demph{39.7} & \demph{-} & \demph{133.3} & \demph{-} & \demph{109.6} & \demph{14.7}  & \demph{-} & \demph{-}  \\
  \demph{LEMON-B~\cite{hu2022scaling}} & \demph{200M} &  \demph{40.3} & \demph{30.2} & \demph{133.3} & \demph{23.3} & \demph{106.8} & \demph{14.1} & \demph{-} & \demph{-} \\
    \demph{SimVLM-B~\cite{wang2021simvlm}} & \demph{1.8B} & \demph{39.0} & \demph{32.9} & \demph{134.8} & \demph{24.0} & \demph{-} & \demph{-} &  \demph{94.8} & \demph{13.1}\\
\midrule
  \ModelName-B & 4M & 39.1 & 30.4 & 128.4 & 23.1 & 88.6 & 13.0 & 86.0 & 12.9 \\
  \ModelName-GOLD-B & 4M & \bf 40.3 & \bf 30.7 & \bf 133.6 & \bf 23.6 & \bf 92.8 & \bf 13.4 & \bf 90.6 & \bf 13.4 \\
  \midrule
  \midrule
  \multicolumn{9}{l}{  \it{Models trained with CIDEr optimization} } \\
     \midrule
  ViTCAP~\cite{fang2022injecting} & 4M & 41.2 & 30.1 & 138.1 & 24.1 & 89.2  & 12.7 & - & - \\
  X-VLM~\cite{zeng2021multi} & 4M  &  41.3 & - & 140.8 & - & - & - & - & -\\
   VinVL-B~\cite{zhang2021vinvl} & 5.7M  & 40.9 & 30.9 & 140.4 & \bf 25.1 & 94.3$^*$ & 13.1$^*$ & 92.5$^*$ & 13.1$^*$\\
   \demph{LEMON-B~\cite{hu2022scaling}} & \demph{200M} & \demph{41.6} & \demph{31.0} & \demph{142.7} & \demph{25.1} & \demph{-} & \demph{-} & \demph{-} & \demph{-} \\
  \midrule
  \ModelName-B & 4M & 42.8 & 31.0 & 142.8 & 24.3 & 96.7 & 13.4 & 94.1 & 13.4 \\
  \ModelName-GOLD-B & 4M & \bf 43.4 & \bf 31.3 & \bf 144.4 &  24.6 & \bf 99.2 & \bf 13.7 & \bf 97.1 & \bf 13.8 \\
  \midrule
  \end{tabular}
  \vspace{-2mm}
  \caption{Results of base-size models on image captioning. We grey models pre-trained on larger magnitudes of data. Numbers with `*' are obtained with constrained beam search during inference and without VLP. The complete results on all metrics are provided in Appendix. B@4: BLEU@4, M: METEOR, C: CIDEr, S: SPICE.
  }
  \label{tab:caption}
   \end{center}
   \vspace{-5mm}
\end{table*}

%% file: tables/results_phrase_grounding.tex
\begin{table}[t]
\setlength{\tabcolsep}{5pt}
\begin{center}
\small
\def \arraystretch{0.98}
\begin{tabular}{cccccccccc}
 \toprule
\multirow{2}{*}{\textbf{Model}} & \multirow{2}{*}{\textbf{\tabincell{c}{Image \\ Backbone}}} &  \multirow{2}{*}{\textbf{\tabincell{c}{\#Pretrain Images \\ (fine-grained)}}}& \multicolumn{3}{c}{ \textbf{Flickr30k Val}} &  \multicolumn{3}{c}{\textbf{Flickr30k Test}} \\
 \cmidrule(lr){4-6} \cmidrule(l){7-9}
        & &  & \textbf{R@1} & \textbf{R@5} &\textbf{ R@10}  & \textbf{R@1} & \textbf{R@5} &\textbf{ R@10} \\
 \midrule
 Visual-BERT~\cite{li2019visualbert} & ResNet-101 & 120k & 70.4 & 84.5 & 86.3 & 71.3 & 85.0 & 86.5 \\
MDETR~\cite{kamath2021mdetr} & EN-B5 &  200k   & 83.6 & 93.4 & 95.1 & 84.3 & 93.9 & 95.8 \\  
GLIP~\cite{li2021grounded} & Swin-B & 860k  & 85.7 & 95.0 & 96.2 & 86.1 & 95.5  &   96.4 \\
\midrule
  \multicolumn{9}{l}{ \demph{ \it{Models pre-trained on more data and/or with larger size} } }\\
  \midrule
\demph{GLIP \cite{li2021grounded}} & \demph{Swin-L} &\demph{ 27M}  & \demph{86.7} & \demph{96.4} & \demph{97.9}    &  \demph{87.1}  & \demph{96.9} &  \demph{98.1} \\
\midrule
\ModelName-B & Swin-B&  860k& \textbf{87.1 }& \textbf{96.1}&  97.4  & \textbf{87.4} & \textbf{96.4} &  97.6   \\
w/o C.G. VLP & Swin-B&  860k & 86.2 & 96.0 & \textbf{97.6} & 86.5 & \textbf{96.4} & \textbf{97.7}   \\
\midrule
\end{tabular}
\caption{Phrase grounding performance on Flickr30k entities dataset. We reproduce GLIP-Base sized results, and GLIP-Large sized results are taken from \cite{li2021grounded}. \ModelName with Base size outperforms a GLIP-L which is trained with 25x more fine-grained data on the R@1 metric. Further, \ModelName without coarse-grained VL pretraining outperforms GLIP-B when trained on the same fine-grained data. 
\label{tab:flickr}}
\end{center}
\vspace{-3mm}
\end{table}

%% file: tables/results_referring_expression.tex
\begin{table*}[t]
\setlength{\tabcolsep}{4pt}
\begin{center}
\small
\resizebox{1.0\textwidth}{!}
{
 \begin{tabular}{ccccccccccc} 
 \toprule
\multirow{2}{*}{\textbf{Model}}  & \multicolumn{2}{c}{\textbf{Pre-training data}} & \multicolumn{3}{c}{\textbf{RefCOCO}} &  \multicolumn{3}{c}{\textbf{RefCOCO+}} & \multicolumn{2}{c}{\textbf{RefCOCOg}}  \\ [0.5ex] 
 \cmidrule(lr){2-3} \cmidrule(lr){4-6} \cmidrule(lr){7-9} \cmidrule(l){10-11}
      &  \textbf{Im-Txt}  & \textbf{Im-Txt-Box} &  \textbf{val} & \textbf{testA} & \textbf{testB} & \textbf{val} & \textbf{testA} & \textbf{testB} &\textbf{val} & \textbf{test}  \\
 \midrule
 
 MDETR-B~\cite{kamath2021mdetr} & & \checkmark & 87.51 & 90.40 & 82.67  & 81.13  & 85.52 &   72.96  &  83.35 & 83.31  \\
UNICORN-B~\cite{yang2021crossing} & & \checkmark & 88.29 & 90.42 & 83.06 & 80.30 & 85.05 & 71.88 & 83.44 & 83.93 \\
\midrule
 \multicolumn{9}{l}{ \demph{ \it{Models pre-trained on more data and/or with larger size} } }\\
  \midrule
\demph{UNITER-L~\cite{chen2020uniter}}   & \demph{\checkmark} & & \demph{81.41} & \demph{87.04} & \demph{74.17} & \demph{75.90} & \demph{81.45} & \demph{66.70} & \demph{74.86} & \demph{75.77}  \\
 \demph{VILLA-L~\cite{gan2020large}} &  \demph{\checkmark} & & \demph{82.39} & \demph{87.48} & \demph{74.84} & \demph{76.17} & \demph{81.54} & \demph{66.84} & \demph{76.18}  & \demph{76.71} \\ 
\demph{OFA-L~\cite{wang2022unifying}}  & \demph{\checkmark}  & \demph{\checkmark} & \demph{90.05} & \demph{92.93} & \demph{85.26}     & \demph{84.49} &   \demph{90.10}  &  \demph{77.77} & \demph{84.54}  & \demph{85.20}  \\
\midrule

\ModelName-B & \checkmark & \checkmark & \textbf{ 90.68} & \textbf{92.59} &  \textbf{87.26}   & \textbf{85.74}  & \textbf{90.13}  & \textbf{ 79.38 }  & \textbf{87.11}  & \textbf{87.32}  \\

 \midrule
\end{tabular}
}
\vspace{-2mm}
\caption{ Results on referring expression comprehension datasets.
}
\label{tab:refexp}
\end{center}
\vspace{-3mm}
\end{table*}

%% file: tables/results_object_detection.tex
\begin{table}[t]
\setlength{\tabcolsep}{3pt}
\begin{center}
\small
\def \arraystretch{0.98}
\begin{tabular}{cccccccc}
 \toprule
\multirow{2}{*}{\textbf{Model}}  &  \textbf{COCO Val 2017} & \multicolumn{4}{c}{\textbf{LVIS MiniVal}} & \multirow{2}{*}{\textbf{ODinW}} \\
\cmidrule(lr){2-2} \cmidrule(lr){3-6} 
    &   \textbf{AP} & \textbf{APr} & \textbf{APc} & \textbf{APf} & \textbf{AP} & \\
 \midrule
     &   Zero-shot/Fine-tune &  \multicolumn{4}{c}{Zero-shot/Fine-tune} & Zero-shot/Fine-tune \\
\midrule
 Mask R-CNN~\cite{he2017mask} &  -  & - /26.3  & - /34.0 & - /33.9 & - /33.3 & - \\
 MDETR~\cite{kamath2021mdetr} & - & - /20.9 & - /24.9 & - /24.3 & - /24.2 & - \\
 GLIP-T~\cite{li2021grounded} & 46.7/55.1 & 17.7/ - & 19.5/ - & 31.0/ - & 24.9/ - & 44.4/63.9\\ 
 GLIP-B~\cite{li2021grounded} &  48.1/57.0  & 17.0/31.3  & 23.9/48.3  & 35.9/56.9  &  29.1/51.0 & 44.8/65.8   \\
 \midrule
 \multicolumn{7}{l}{ \demph{ \it{Models pre-trained on more data and/or with larger size} } }\\
  \midrule
 \demph{GLIP-L~\cite{li2021grounded}} & \demph{49.8/60.8}  & \demph{28.2/ - } & \demph{34.3/ - } & \demph{41.5/ - } & \demph{37.3/ - } & \demph{52.1/68.9} \\
\midrule
 \ModelName-B &   \textbf{49.3/58.4} & \textbf{29.5/50.0}  & \textbf{32.2/56.9}  & \textbf{40.1/58.1} &  \textbf{35.8/56.9 }& \textbf{47.0/65.9}  \\
 \midrule
\end{tabular}
\caption{ Zero-shot transfer and fine-tuning results for object detection on COCO, LVIS and the average over 13 datasets for object detection in the wild. Detailed scores on the 13 datasets are presented in the Appendix. \ModelName achieves better AP across the board compared to similarly sized GLIP-B, trained on the same amount of fine-grained data.   On rare objects in LVIS, \ModelName outperforms GLIP-L trained on 25x more fine-grained data. Results without coarse-grained pre-training are provided in the Appendix.
\label{tab:ZS/finetuned_OD}}
\end{center}
\vspace{-3mm}
\end{table}

%% file: subtex/5_conclusions.tex
\section{Conclusion}
\label{sec:limitations}

We propose ($i$) \ModelName, a novel architecture and ($ii$) a coarse-to-fine pre-training pipeline.
We perform extensive experiments and show consistent improvements over strong baselines across a diverse set of tasks. The results demonstrate the effectiveness of \ModelName coupled with our pre-training strategy, by setting new SoTA scores while at the same time reducing the requirement of expensive box-level annotations. Future directions include scaling our models and extending our framework to other modalities.

The approach introduced in our work can potentially inherit undesirable societal biases that exist in our pre-training data. Careful debiasing and filtering of data should be undertaken before real-life deployment of our work. Additionally, pre-training can induce environmental costs, and minimizing these costs is an avenue that we plan to explore further. 

\section*{Acknowledgement}
We would like to thank Nguyen Bach, Jiayuan Huang, and Luis Vargas for their support. We also thank Wenhui Wang, Li Dong, Furu Wei, Bin Xiao, and Lu Yuan for their helpful discussions. We also thank Liunian Harold Li and Te-Lin Wu for their feedback on the manuscript.
Aishwarya is supported in part by the National Science Foundation under NSF Award 1922658. Zi-Yi is supported in part by the DARPA Machine Common Sense (MCS) program under Cooperative Agreement N66001-19-2-4032 and NIH R01HL152270.

%% file: subtex/6_supp.tex
\appendix
\section{Appendix}
\subsection{Implementation Details}

\paragraph{Vision-Language Classification.} For the VL classification tasks, we follow METER~\cite{dou2021empirical} to set the hyper-parameters. Specifically, we fine-tune our models with the peak learning rates of 2e-5 for the backbones, 1e-4 for the cross-modal parameters, and 1e-3 for the head layer for 10 epochs. The batch size is set to 512. The image resolutions are set to 576 for VQAv2 and 384 for NLVR$^2$. We evaluate models with the VQA scores for VQAv2 and accuracy for NLVR$^2$.  RandAugment~\cite{cubuk2020randaugment} is used during the downstream fine-tuning stage.

\paragraph{Image-Text Retrieval.} For image-text retrieval, we remove the cross-attention layers in the backbones and use the dual encoder architecture. We set the peak learning rates to 2e-5 for the backbones and 1e-4 for the head layer. The batch size is set to 1024. The image resolutions are set to 576 for both COCO and Flickr30k. We evaluate on the Recall@1,5,10 metrics for both text and image retrieval.

\paragraph{Image Captioning.} For image captioning, we only keep the image-to-text attentions and feed the image representations in the last layer of the image encoder to the cross-attention modules. In this way, the model is turned into a standard seq2seq model, and we use the causal mask in the decoding side and predict outputs auto-regressively. We first train our models with the cross-entropy loss for 5 epochs with the peak learning rates of 5e-5 for the backbones, and 2.5e-4 for the rest of the parameters. Then, we fine-tune it with GOLD~\cite{pang2021text} for 5 epochs as it is efficient and has proven to be effective when the model input can correspond to different outputs. We set the peak learning rate to 1e-5 for the backbones during GOLD training. For CIDEr optimization, the learning rate is further reduced to 1e-6 and we train the models for 3 epochs. The batch size is set to 512. We use a beam size of 5 during inference and do not use constrained beam search. We use the same model when testing on COCO and NoCaps, and we evaluate on BLEU~\cite{papineni2002bleu}, METEOR~\cite{banerjee2005meteor}, CIDEr~\cite{vedantam2015cider}, and SPICE~\cite{anderson2016spice} metrics.

\paragraph{Phrase Grounding.}
For phrase grounding on Flickr30k, we do not further fine-tune the model after fine-grained pre-training, and just directly evaluate on the Recall@ 1,5,10 metrics. 

\paragraph{Referring Expression Comprehension (REC).}
For the REC datasets, we use batch size 16 and fine-tune on the respective dataset for 20 epochs. We use a warmup of 2000 steps, with a peak learning rate of 1e-5 for both the OD head as well as the rest of the model's paramaters, with two learning rate drops at 67\% and 89\% of the total number of steps. We switch off the horizontal flip augmentation during REC training, as we find that it adversely affects the performance, especially on the RefCOCO dataset, which includes many examples having degenerate language such as just ``left'' or ``right'' rather than using descriptive words for the referring expressions.

\paragraph{Object Detection.}
For both COCO and LVIS detection, we train for 24 epochs, with batch size 32, with a learning rate of 1e-5 for the whole model,  with two learning rate drops at 67\% and 89\% of the total number of steps. For the ODinW datasets, we fine-tune for 12 epochs, with early stopping based on the validation accuracy. 

The object detection data is constructed as follows - The object category names are directly used in their text form separated by full stops as input to the text encoder. We follow the same protocol as in GLIP [3] to be comparable to their experiments. More specifically, the input text will look like this: "person. bicycle. car. .... toothbrush", and the model will learn how to ground image regions into these object names. An example of input and output predicted by the model can be seen in Fig. \ref{fig:Finegrained_viz}.

\subsection{Ablation Study}\label{sec:ablation}

\begin{figure*}
  \centering
\begin{subfigure}{0.32\linewidth}
    \includegraphics[width=1.0\linewidth]{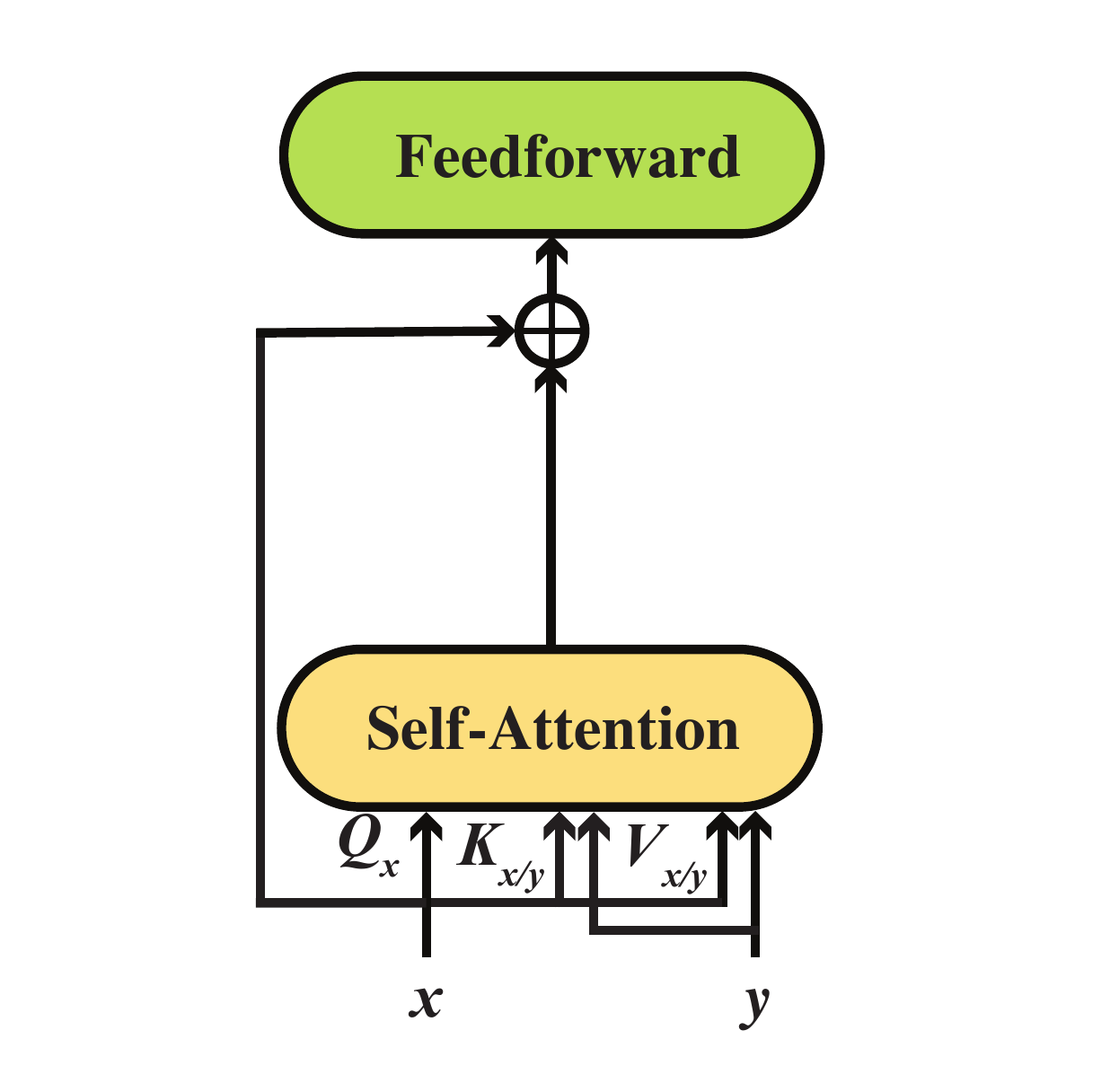}
    \caption{merged attention}
    \label{fig:archa}
  \end{subfigure}
  \begin{subfigure}{0.32\linewidth}
    \includegraphics[width=1.0\linewidth]{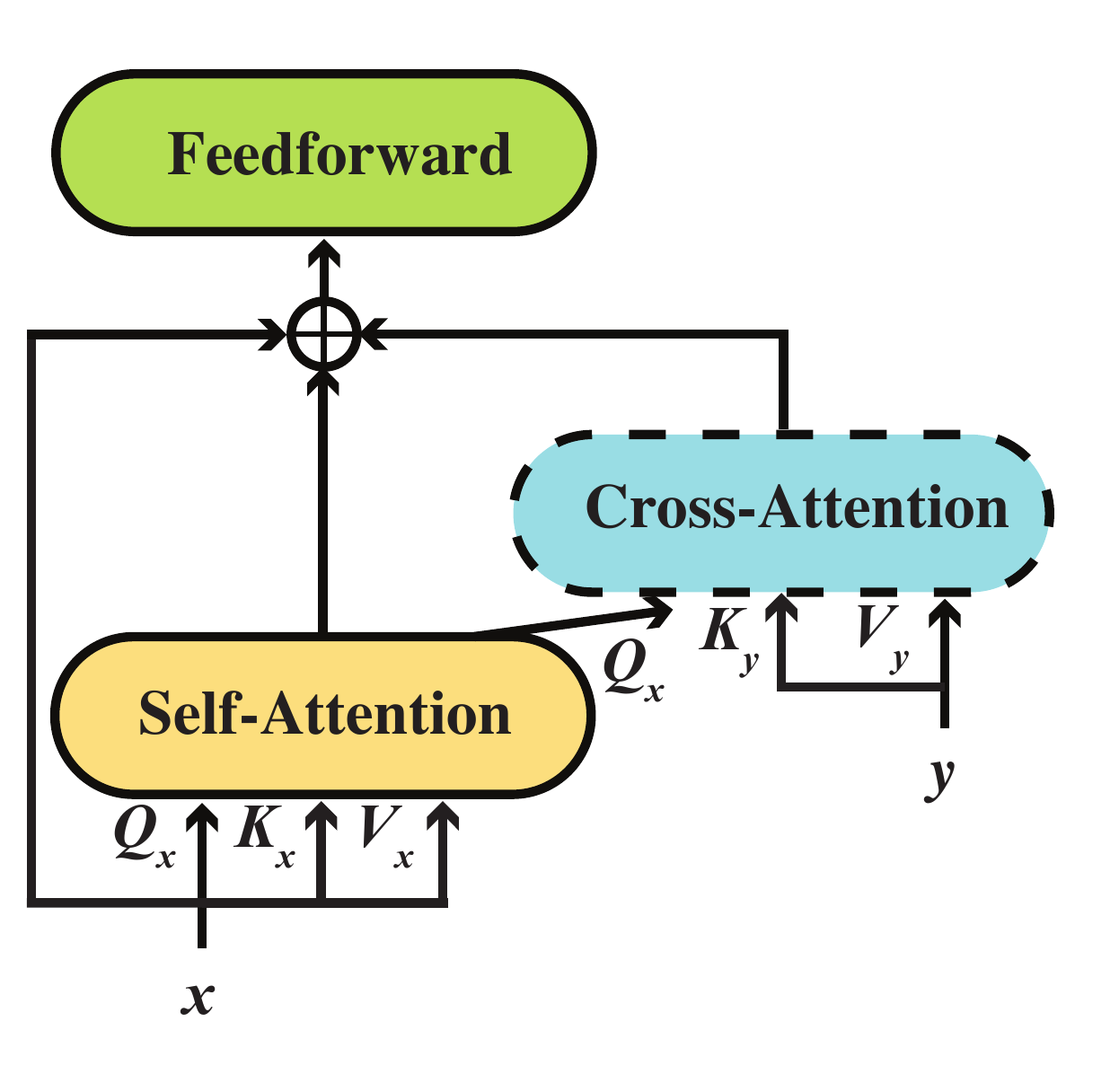}
    \caption{co-attention w/o $\alpha$}
    \label{fig:archb}
  \end{subfigure}
  \begin{subfigure}{0.32\linewidth}
    \includegraphics[width=1.0\linewidth]{figures/fib-arch4.pdf}
    \caption{co-attention w/ $\alpha$}
    \label{fig:archc}
  \end{subfigure}
  \caption{Different strategies for fusion in the backbone. ($\xv$, $\yv$) are the (image, text) or (text, image) representations, and $\alpha$ is a learnable scalar.
  \label{fig:arch}}
\end{figure*}

\input{supp_tables/ablation_arch}

\paragraph{Ablation Study on the Fusion Strategies.}
We perform ablation studies on our fusion module. We investigate three different fusion strategies as shown in Figure~\ref{fig:arch}. Merged attention concatenates representations from the two input modalities and feeds them into the self-attention layer for fusion. Note that here the key and value matrices for the two modalities are different. On the other hand, co-attention inserts a cross-attention layer into each of the encoding layer. The insertion of the cross-attention layer offers the flexibility of controlling to what extent we want the two modalities to fuse together as we can easily introduce an $\alpha$ term into the module as in Figure~\ref{fig:arch-ours}.

As shown in Table~\ref{tab:ablation_arch}, we compare the three fusion strategies by directly fine-tuning our models without performing VLP for efficiency. We use Swin Transformer and RoBERTa as our vision and text backbones and load their pre-trained parameters for initialization. We set the image resolution to 224$\times$224. We can see that merged attention and co-attention achieve comparable performance without $\alpha$. For both strategies, increasing the number of fusion layers can lead to performance drop. However, after introducing $\alpha$, we can see significant improvements of co-attention, indicating the importance of having an explicit controlling/gating mechanism for fusion in the backbone. 

After the $\alpha$ term is introduced, we can increase the number of fusion layers and achieve robust performance. Based on the ablation results, we choose to fuse the top 6 layers of the backbones as it can achieve a good accuracy-efficiency trade-off.

\input{supp_tables/ablation_objectives}
\paragraph{Ablation Study on Pre-training Objectives.} 
Following previous work~\cite{li2021align,dou2021empirical,wang2021vlmo}, we pre-train our models with image conditioned masked language modeling, image-text matching with hard negative mining, and image-text contrastive losses during the coarse-grained pre-training stage. In this part, we ablate each of the pre-training objectives and evaluate our models on both VQAv2 and Flickr30k retrieval tasks. Specifically, we use Swin Transformer and RoBERTa as our vision and text backbones and load their pre-trained parameters for initialization. The image resolution is set to 224$\times$224 and we pre-train models for 100k steps with 1,024 batch size. We use AdamW with the peak learning rates of 1e-4 for the backbones and 5e-4 for the cross-modal parameters. We use linear warmup over the first 1k steps and linear decay.

As shown in Table~\ref{tab:ablation_obj}, we can see that removing any of the pre-training objectives can lead to performance drop, and hard negative mining can bring improvements on both VQA and retrieval tasks. Masked language modeling is most effective for VQA, while removing it will not hurt the retrieval performance. This set of experiments demonstrates that all of the objectives are necessary for our models to obtain good performance.

\input{supp_tables/ablation_coarse.tex}
\paragraph{Ablation Study on the Two-Stage Pre-training.} In this paper, we propose a coarse-to-fine pre-training strategy for handling VL tasks of different kinds. In this paragraph, we remove the coarse-grained pre-training stage and only pre-train the models with image-text-box data and see how it performs. As shown in Table~\ref{tab:ablation_coarse}, we see gains across both tasks when utilizing the coarse-grained pre-training. Similar to the case of Flickr30k, on RefCOCO+ the coarse-grained pre-training helps FIBER to get better performance than large-sized model trained with more data. In addition, note that without the coarse-grained pre-training, the only difference between FIBER and GLIP is the architectural difference, and the fact that FIBER can still outperform GLIP in this setting demonstrates the effectiveness of our proposed architecture.

\input{supp_tables/ablation_back.tex}
\paragraph{Ablation Study on Different Backbones.} While previous work~\cite{dou2021empirical} has compared different vision and text backbones for VLP models, we investigate if their conclusions stil apply in our settings. Specifically, we try BERT and RoBERTa for our text encoder and CLIP-ViT and Swin Transformer for our image encoder. As shown in Table~\ref{tab:ablation_back}, we can see that RoBERTa and Swin Transformer perform slightly better than BERT and CLIP-ViT before VLP, which is consistent with previous findings in METER~\cite{dou2021empirical}. Note that while CLIP-ViT has the potential to perform better than Swin Transformer after VLP, it is hard to be adapted for region-level tasks such as object detection. Therefore, pairing Swin Transformer with RoBERTa is the optimal configuration in our settings.

\begin{figure*}
  \centering
  \begin{subfigure}{0.4\linewidth}
    \includegraphics[width=1.0\linewidth]{figures/taskc.pdf}
    \caption{seq2seq}
    \label{fig:task_seq2seq}
  \end{subfigure}
  \begin{subfigure}{0.4\linewidth}
    \includegraphics[width=1.0\linewidth]{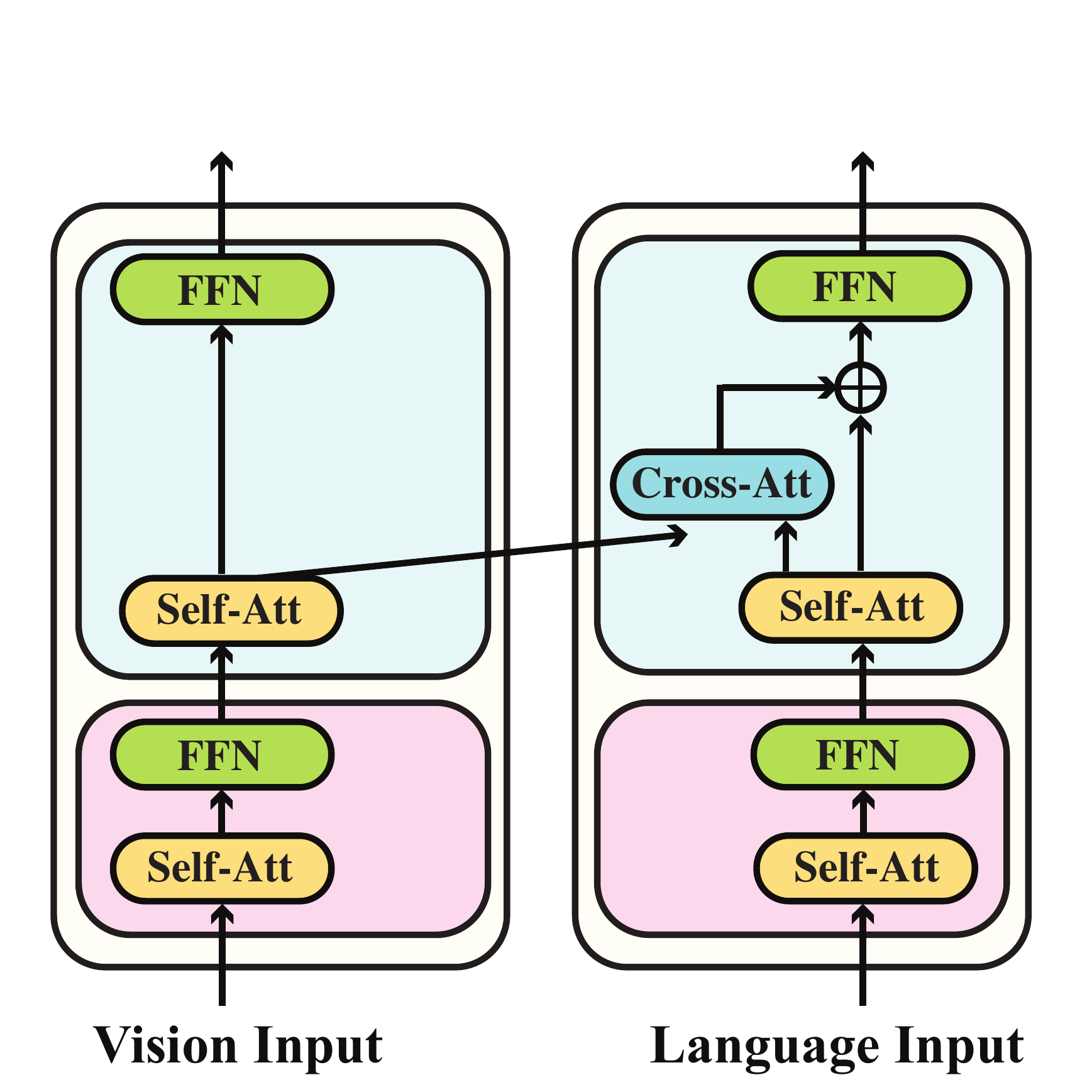}
    \caption{ladder}
    \label{fig:task_ladder}
  \end{subfigure}
  \caption{When adapting \ModelName to image captioning, we can either use the seq2seq structure or the ladder architecture as in pre-training. 
  \label{fig:FIBER_for_captioning}}
\end{figure*}

\subsection{Additional Results}\label{sec:additional_results}

\paragraph{Additional Results on Image Captioning.}
For image captioning, we evaluate the models with BLEU-4, METEOR, ROUGE-L, CIDEr, and SPICE metrics on COCO and NoCaps. On NoCaps, we have fine-grained evaluation results on different domains, including in-domain, near-domain, out-domain, and entire domain settings. In this part, we provide the complete evaluation results in Table~\ref{tab:caption_full_coco},~\ref{tab:caption_full_nocaps_val} and~\ref{tab:caption_full_nocaps_test}. We can see that both GOLD and CIDEr optimization can improve the model performance across metrics. We also see a noticeable performance drop when evaluating our models on out-of-domain data, but complementary methods such as constrained beam search can be used to alleviate the issue. Also, training our models with more captioning data should also be helpful in these settings.

\input{supp_tables/results_captioning_full}

Also, in the main paper, we adapt \ModelName for image captioning by turning it into a standard seq2seq model as in Figure~\ref{fig:task_seq2seq}, where the output of the final encoding layer will be fed into the image-to-text cross-attention modules. Another possible design is to keep the ladder structure as we used in pre-training (Figure~\ref{fig:task_ladder}), so there can be less mismatching between pre-training and fine-tuning. As shown in Table~\ref{tab:caption_full_coco}, the two architectures can achieve comparable performance. Considering that the seq2seq architecture is more widely adopted in the current literature, we decide to use the seq2seq architecture for image captioning.

\input{supp_tables/open_vqa}
\paragraph{Open-ended VQA.}
In most existing literature, VQA is treated as a classification task, where a vocabulary of some most frequent answers are constructed and VL models predict which answer corresponds to the given question based the constructed vocabulary. However, question answering is inherently open-ended. Since we can turn our models into a generative model by fine-tuning on image captioning, we also investigate if our models can perform open-ended VQA in this part. 

Following~\cite{cho2021unifying}, we break down VQA questions into in-domain and out-of-domain questions, where the answers to the out-of-domain questions do not appear in the top-$k$ ($k=3,129$) candidates. We use the Karpathy split~\cite{karpathy2015deep} in this setting.

As shown in Table~\ref{tab:open_vqa}, our generative model can perform better than VL-T5 and VL-BART, while lagging behind SimVLM especially in out-of-domain settings, possibly because SimVLM is trained with over a billion image-caption pairs and is more robust in this setting. The results indicate that our model can be turned into a general open-ended VQA model as well.

\paragraph{Uni-modal Performance.}

It can be interesting to see whether our backbones can still perform uni-modal tasks after VLP. Therefore, in this part, we also evaluate our language backbones on uni-modal tasks. 

\input{supp_tables/uni_modal}

\input{supp_tables/ODinW}
Specifically, we test our language backbone after the first coarse-grained pre-training stage on the GLUE~\cite{wang2018glue} benchmark. As shown in Table~\ref{tab:glue}, the uni-modal performance of our text encoder can drop marginally on some tasks, possibly because the model only encounters simple text captions during VLP. However, it is still better than SimVLM which is trained with 800GB of web-crawled documents from scratch.

{On the other hand, as shown in Table~\ref{tab:image_encoder}, our image encoder can achieve comparable and sometimes even better performance on vision-only tasks including image classification and semantic segmentation after coarse-grained pre-training. The results suggest that the image encoder can remain powerful even if the fusion modules are removed.}

\paragraph{Using the Model Checkpoint After Fine-grained Pre-training for VQA.}
We also test what if we fine-tune the fine-grained pre-trained checkpoint on VQA in this part. We find that after the second-stage fine-grained pre-training, the model performance on the VQAv2 test-dev set can drop from 78.55 to 74.3, indicating that the two-stage pre-training paradigm is indeed necessary for different VL tasks, as tasks of different characteristics can require checkpoints from different pre-training stages.

\paragraph{Detailed Results on ODinW.} Detailed results on the 13 ODinW datasets are provided in Table~\ref{tab:odinw_fullres}.

\subsection{Visualization after coarse-grained pre-training}
\begin{figure*}[h]
  \centering
    \includegraphics[width=1.0\linewidth]{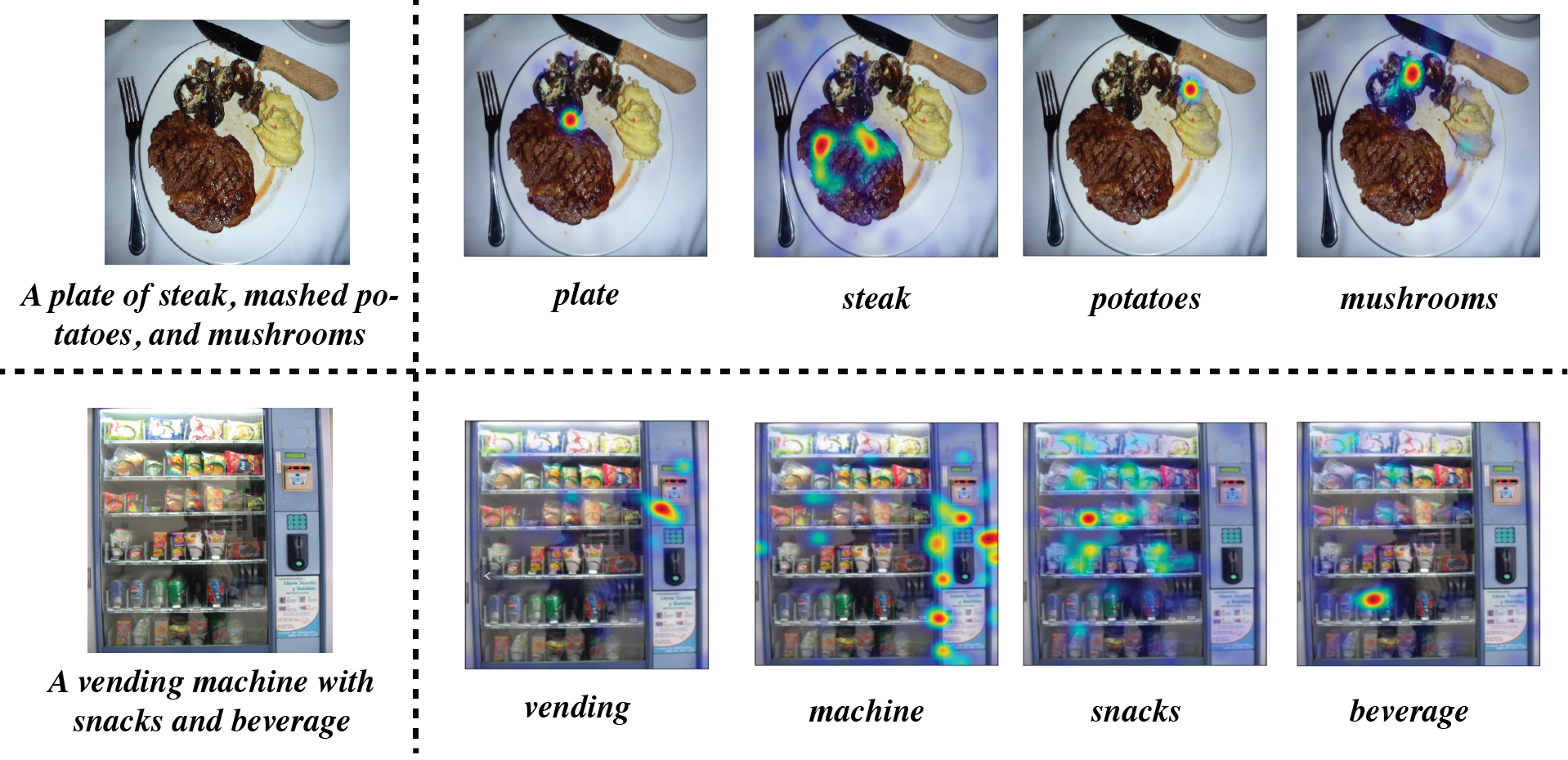}
  \caption{Visualizations of the cross-attention maps obtained by Grad-CAM~\cite{selvaraju2017grad}. Given each of the tokens in a caption, the model can attend to its corresponding regions. The figures are from the NoCaps validation set (ID: 253, 3766).}
  \label{fig:vis}
  \vspace{-3mm}
\end{figure*}

We also provide a qualitative analysis of our model. As shown in Figure~\ref{fig:vis}, we use Grad-CAM~\cite{selvaraju2017grad} to visualize the cross-attention maps of our coarse-grained pre-trained checkpoint. We find that the model can correctly align concepts and image regions for some examples, suggesting that the model can learn visual grounding implicitly.

\subsection{Visualization after fine-grained pre-training }
\begin{figure*}[h]
    \centering
    \includegraphics[width=0.8\linewidth]{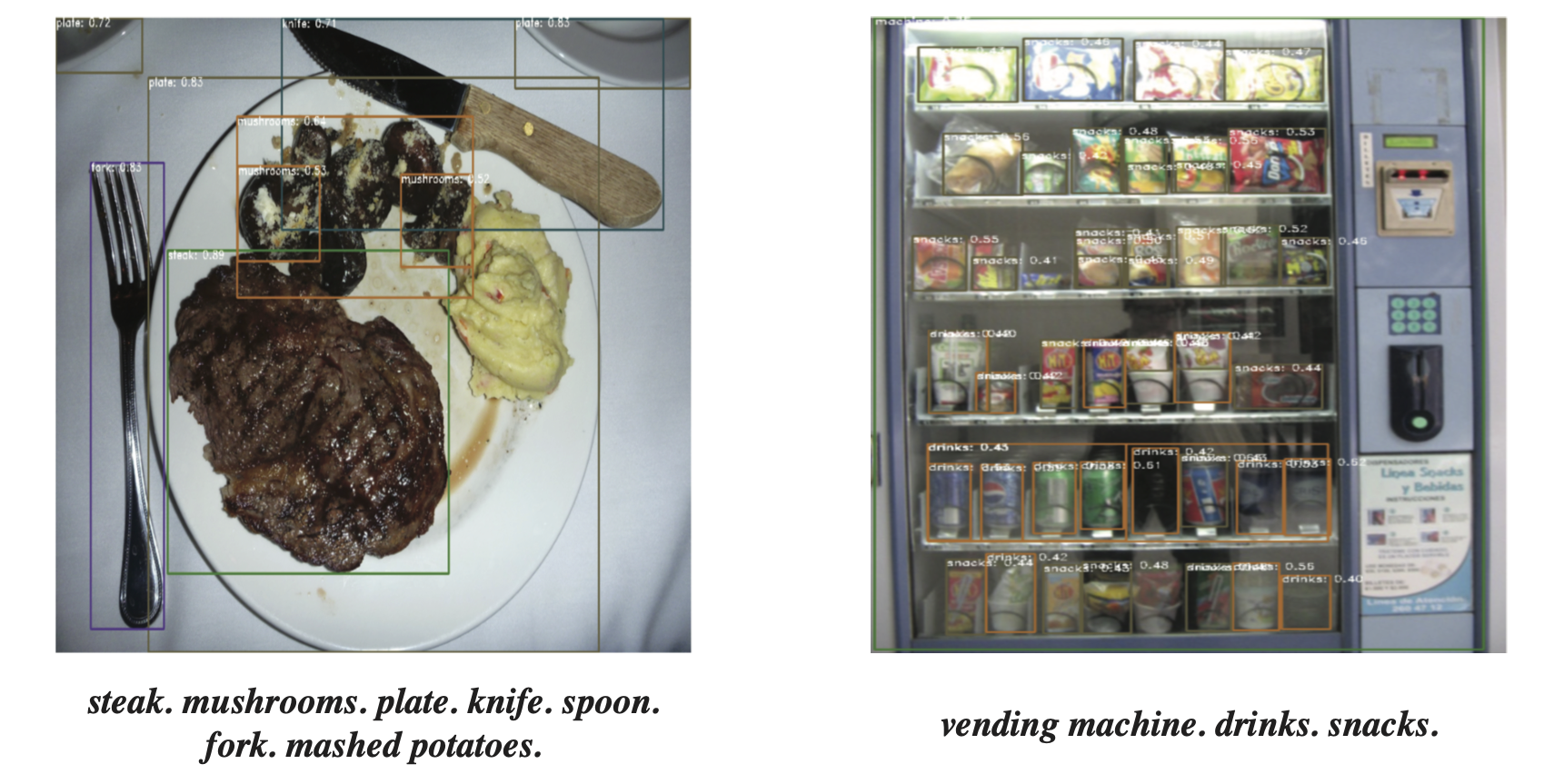}
    \caption{The same images probed after fine-grained pre-training.}
    \label{fig:Finegrained_viz}
\end{figure*}

\begin{figure*}
\centering
\includegraphics[width=0.9\linewidth]{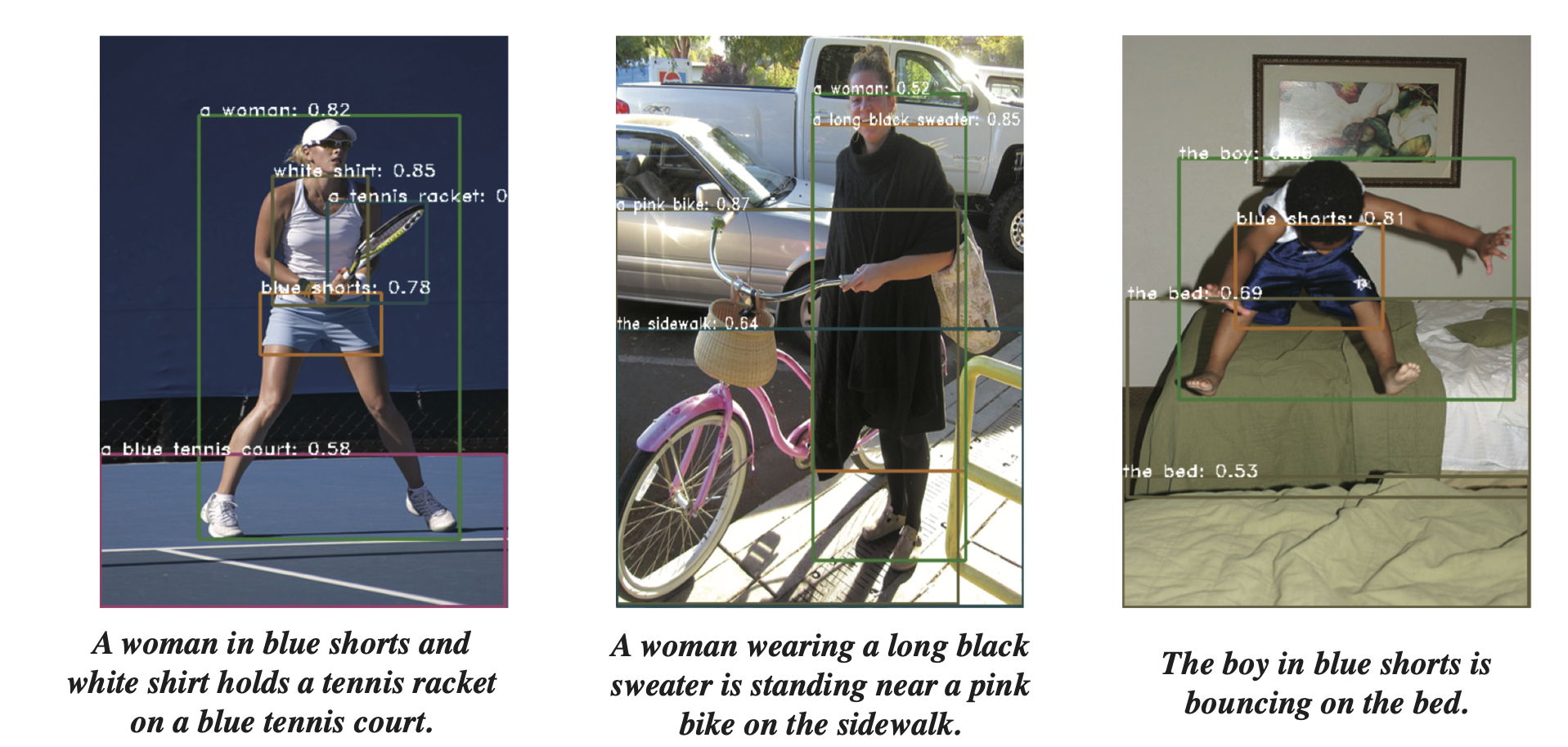}
        \caption{Some examples of phrase grounding from the validation set for Flickr30k entities.}
        \label{fig:Finegrained_viz2}
\end{figure*}

\begin{figure*}
\centering
\includegraphics[width=0.9\linewidth]{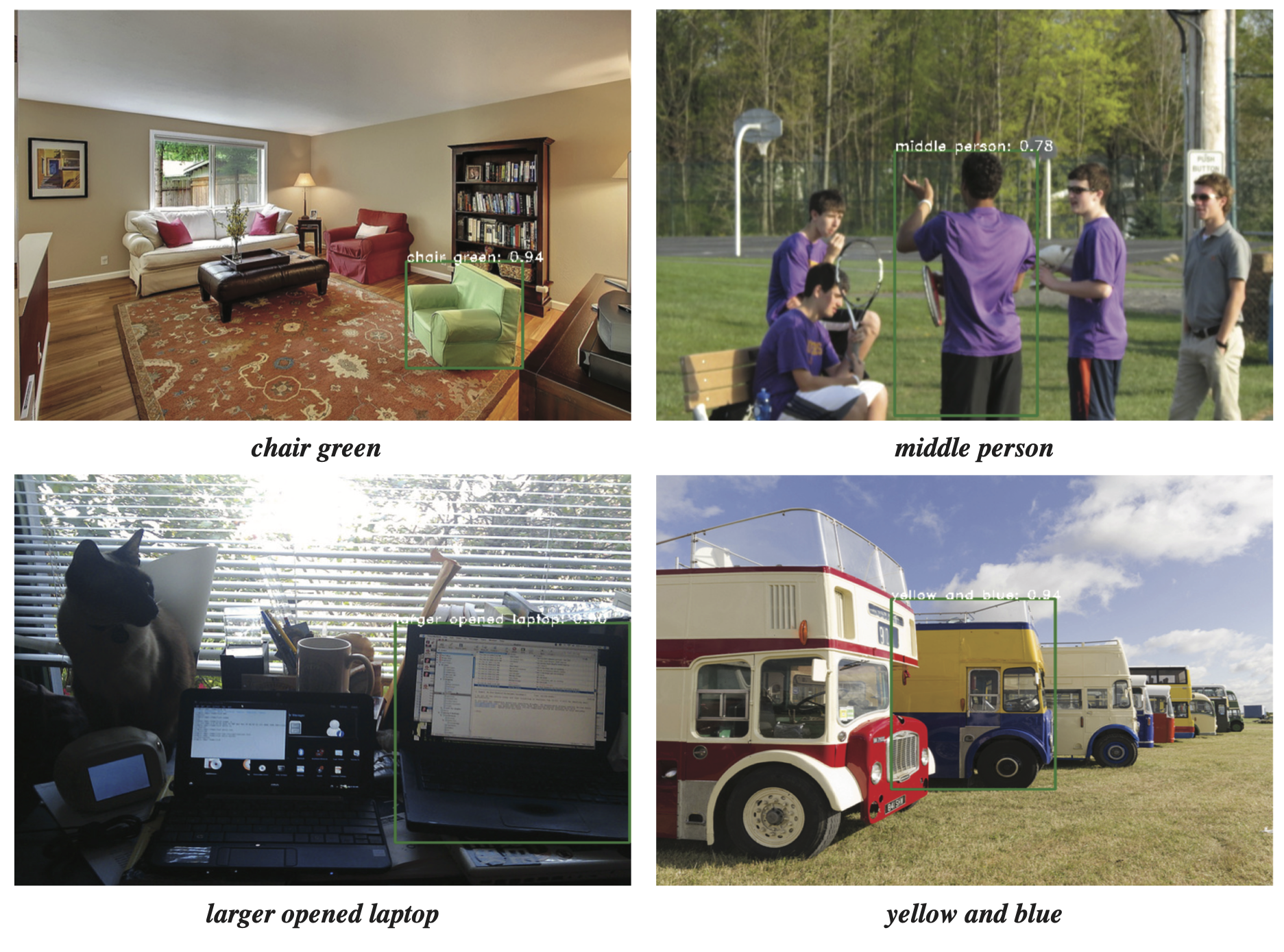}
        \caption{Some examples of referring expression comprehension from the validation set of RefCOCO+. }
        \label{fig:refexp_viz}
        
\end{figure*}

\begin{figure*}
    \centering
    \includegraphics[width=0.9\linewidth]{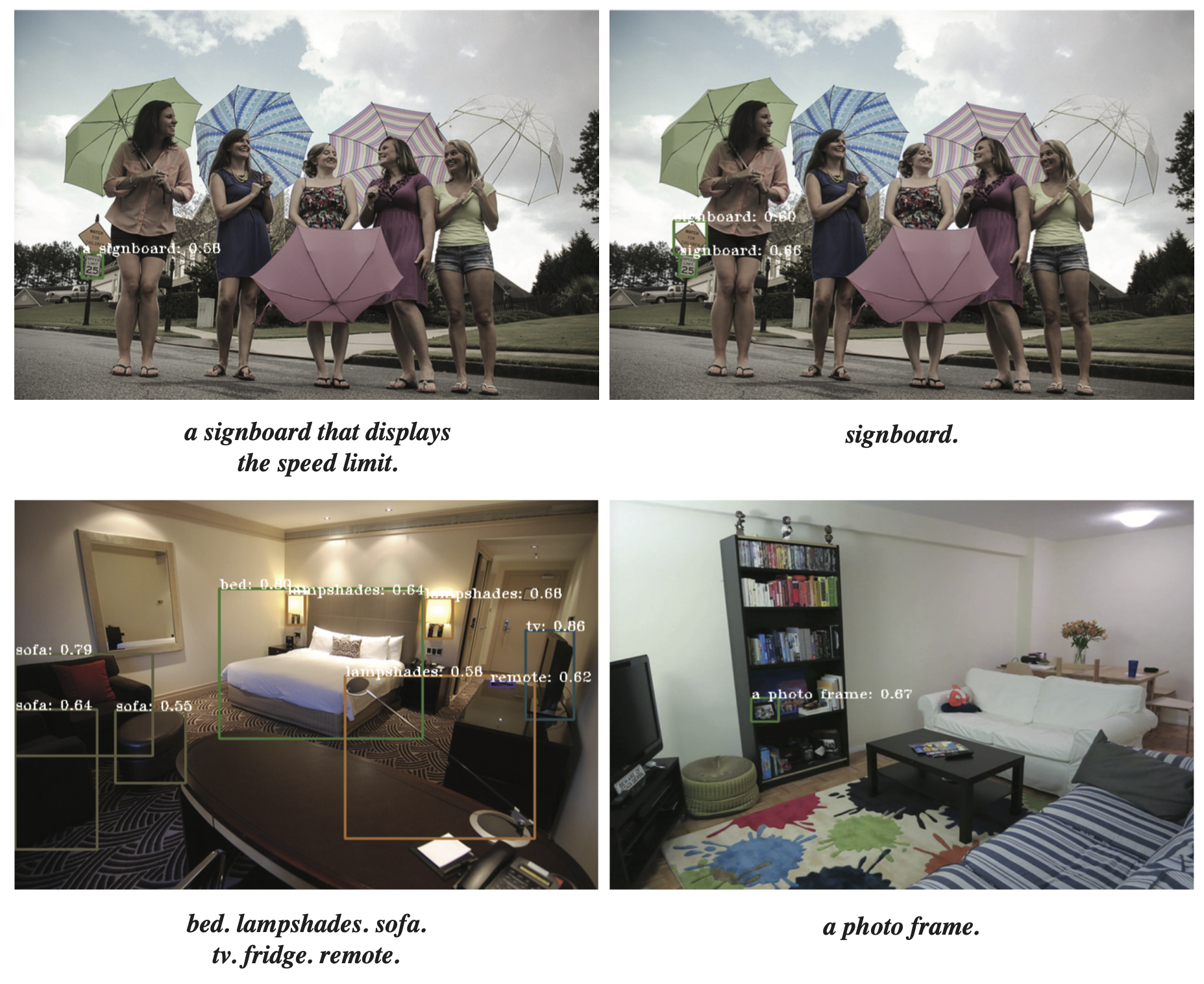}
    \caption{Some images with prompts for various items in the scene.}
    \label{fig:Finegrained_viz3}
\end{figure*}

We also provide visualization after fine-grained pre-training in Figure~\ref{fig:Finegrained_viz},~\ref{fig:Finegrained_viz2},~\ref{fig:refexp_viz}, and~\ref{fig:Finegrained_viz3}.

\subsection{Text and Vision Backbones of Different Models}
In this section, we list the backbones of different models in Table~\ref{tab:model_backbone}.

\begin{table}[h]
\small
\begin{center}
  \begin{tabular}{cccccccccc}
  \toprule
    \bf Model &    \bf Vision Encoder  &  \bf Text Encoder \\
    \midrule
     UNITER & Frozen ResNet-based OD & XFM (init w/ BERT) \\
 VILLA & Frozen ResNet-based OD & XFM (init w/ BERT) \\
 UNIMO & Frozen ResNet-based OD &  XFM (init w/ RoBERTa)\\
 VinVL & Frozen ResNet-based OD & XFM (init w/ BERT) \\
ViLT & XFM (init 5w/ ImageNet-ViT) & XFM (init w/ ImageNet-ViT and BERT embeddings) \\
ALBEF & XFM (init w/ ImageNet-ViT)  & XFM (init w/ BERT) \\
VLMo &  XFM (init w/ ImageNet-ViT) & XFM (init w/ ImageNet-Language-ViT) \\
UFO & XFM (init w/ ImageNet-ViT) & XFM (init w/ ImageNet-ViT) \\
 ViTCAP & XFM (init w/ ImageNet-ViT) & XFM (init w/ ImageNet-ViT and BERT embeddings) \\
METER-Swin & Swin (init w/ ImageNet-Swin) & XFM (init w/ RoBERTa)\\
METER-CLIP & XFM (init w/ CLIP-ViT) & XFM (init w/ RoBERTa) \\
MDETR & EfficientNet& XFM (init w/ RoBERTa) \\
GLIP & Swin (init w/ ImageNet-Swin) & XFM (init w/ BERT) \\
FIBER & Swin (init w/ ImageNet-Swin) & XFM (init w/ RoBERTa)  \\
    \midrule
  \end{tabular}
  \caption{Backbones of different models. XFM stands for Transformer}
  \label{tab:model_backbone}
     \end{center}
   \vspace{-3mm}
\end{table}

%% file: supp_tables/ablation_arch.tex
\begin{table}[t!]
\small
\begin{center}
  \begin{tabular}{cccccccccc}
  \toprule
    \bf Model  &  \bf \#Fusion Layers  &  \bf \#Fusion Params. & \bf VQAv2 \\
    \midrule
    No Fusion & 0 & 0M & 65.65 & \\
    \midrule
    \multirow{2}{*}{Merged Attention} & 3 & 7.9M & 71.24 &  \\
    & 6 & 12.6M & 70.67 &  \\
    \midrule
     \multirow{2}{*}{Co-attention w/o $\alpha$} & 3 & 16.1M & 70.84 &  \\
    & 6 & 26.0M & 68.13 &  \\
    \midrule
     \multirow{4}{*}{Co-attention w/ $\alpha$} & 3 & 16.1M & 71.20 &  \\
    & 6 & 26.0M & 71.97 &  \\
    & 9 & 35.8M & 72.10 &  \\
    & 12 & 45.6M & 72.08 &  \\
    \midrule
  \end{tabular}
  \caption{Ablation study on the fusion strategies. Results are obtained by directly fine-tuning models initialized with uni-modally pre-trained parameters and without VLP. Results on VQAv2 are on test-dev set. }
  \label{tab:ablation_arch}
     \end{center}
   \vspace{-3mm}
\end{table}

%% file: supp_tables/ablation_objectives.tex
\begin{table}[t!]
\small
\begin{center}
\def \arraystretch{0.98}
  \begin{tabular}{cccccccccccc}
    \toprule
  \multicolumn{4}{c}{\bf Pre-training Objectives} &  \multicolumn{1}{c}{\bf VQAv2} & \multicolumn{2}{c}{\bf Flickr30k} \\
  \cmidrule(lr){1-4} \cmidrule(lr){5-5} \cmidrule(lr){6-7}   
   \bf MLM & \bf ITM & \bf ITM-hard  & \bf ITC & \bf test-dev & \bf IR@1 & \bf TR@1 \\
    \midrule
    \checkmark & \checkmark & \ding{53} & \ding{53} & 72.47 & 65.50 & 79.30 \\
    \checkmark  & \ding{53} & \ding{53} & \checkmark & 74.16 & 73.74 & 87.70 \\
    \ding{53} & \ding{53} & \checkmark & \checkmark & 67.45 & 75.20 & 87.00 \\
    \checkmark & \checkmark & \ding{53} & \checkmark & 74.49 & 73.58 & 87.80 \\
    \checkmark & \ding{53} & \checkmark & \checkmark & 75.98 & 75.26 & 87.50 \\
  \midrule
\end{tabular}
\caption{Ablation study on the pre-training objectives and whether the hard negative mining strategy is necessary in the coarse-grained pre-training stage. }
 \label{tab:ablation_obj}
\end{center}
\vspace{-8mm}
\end{table}

%% file: supp_tables/ablation_coarse.tex
\begin{table}[t!]
\small
\begin{center}
  \begin{tabular}{cccccccccc}
  \toprule
       \bf  Model & \bf OD on COCO & \bf OD on LVIS & \bf ODinW & \bf RefCOCO+ \\
    \midrule
    & Zero-shot/Fine-tune &  Zero-shot/Fine-tune &  Zero-shot & Val/TestA/TestB \\
    \midrule
   OFA-L & - & - & - & 84.49/90.10/ 77.77 \\
   GLIP-B	& 48.1/57.0	& 29.1/51.0	 & 44.8	 & - \\
   FIBER-B w/o C.G. VLP &	48.9/57.8	& 31.6/55.8	& 45.1	 &85.04/88.82/78.59 \\
   FIBER-B &	49.3/58.4 &	35.8/56.9	& 47.0	& 85.74/90.13/79.38 \\
    \midrule
  \end{tabular}
  \caption{Ablation study on our proposed two-stage pre-training strategy.}
  \label{tab:ablation_coarse}
     \end{center}
   \vspace{-3mm}
\end{table}

%% file: supp_tables/ablation_back.tex
\begin{table}[t!]
\small
\begin{center}
  \begin{tabular}{cccccccccc}
  \toprule
       \bf Vision Encoder  &  \bf Text Encoder & \bf VQAv2 \\
    \midrule
   Swin & RoBERTa & 71.97 \\
   Swin & BERT & 71.86 \\
   CLIP-ViT & RoBERTa & 71.37 \\
    \midrule
  \end{tabular}
  \caption{Results of different vision and text backbones for FIBER without VLP.}
  \label{tab:ablation_back}
     \end{center}
   \vspace{-3mm}
\end{table}

%% file: supp_tables/results_captioning_full.tex
\begin{table*}[t]
\setlength{\tabcolsep}{6pt}
\begin{center}
\small
\def \arraystretch{0.95}
 \begin{tabular}{ccccccccccccccc} 
 \toprule
    \multirow{2}{*}{\bf Model}  & \multicolumn{5}{c}{\bf COCO}  \\
    \cmidrule(lr){2-6} 
     & \bf BLEU@4 & \bf METEOR & \bf ROUGE-L & \bf CIDEr &\bf SPICE  \\
    \midrule
     \multicolumn{6}{l}{  \it{Models trained without CIDEr optimization} } \\
     \midrule
      \ModelName-Ladder-B & 38.6 & 30.1 & 58.8 & 127.5 & 22.8 \\
  \ModelName-B &  39.1 & 30.4 & 59.3 & 128.4 & 23.1   \\
  \ModelName-GOLD-B &   40.3 & 30.7 & 60.0 &  133.6 & 23.6   \\
  \midrule
  \midrule
  \multicolumn{6}{l}{  \it{Models trained with CIDEr optimization} } \\
     \midrule
  \midrule
  \ModelName-B &  42.8 & 31.0 & 61.5 & 142.8 & 24.3   \\
  \ModelName-GOLD-B &  43.4 & 31.3 & 61.8 & 144.4 &  24.6   \\
  \midrule
  \end{tabular}
  \vspace{-2mm}
  \caption{The complete set of results on COCO image captioning, with another model variant \ModelName-Ladder. See Figure~\ref{fig:FIBER_for_captioning} for details.
  }
  \label{tab:caption_full_coco}
   \end{center}
   \vspace{-4mm}
\end{table*}

\begin{table*}[t]
\setlength{\tabcolsep}{3pt}
\begin{center}
\resizebox{\textwidth}{!}
{
 \begin{tabular}{cccccccccccccccccccccccccccccccccccccc} 
 \toprule
    \multirow{2}{*}{\bf Model}  & \multicolumn{5}{c}{\bf in-domain} & \multicolumn{5}{c}{\bf near-domain} & \multicolumn{5}{c}{\bf out-domain} & \multicolumn{5}{c}{\bf entire}  \\
    \cmidrule(lr){2-6}    \cmidrule(lr){7-11}    \cmidrule(lr){12-16}    \cmidrule(lr){17-21} 
     & \bf B@4 & \bf M & \bf R & \bf C &\bf S   & \bf B@4 & \bf M & \bf R & \bf C &\bf S   & \bf B@4 & \bf M & \bf R & \bf C &\bf S  & \bf B@4 & \bf M & \bf R & \bf C &\bf S  \\
    \midrule
     \multicolumn{6}{l}{  \it{Models trained without CIDEr optimization} } \\
     \midrule
  \ModelName-B & 29.7 & 30.0 & 58.2 & 98.5 & 13.9 & 24.4 & 27.5 & 55.6 & 88.2 & 13.0 & 18.0 & 25.4 & 53.5 & 82.8 & 12.2 &  23.9 & 27.5 & 55.6 & 88.6 & 13.0 \\
  \ModelName-GOLD-B &  29.7 & 30.1 & 58.2 & 100.6 & 14.0 & 26.8 & 28.2 & 57.0 & 92.9 & 13.5 & 18.3 & 25.8 & 54.3 & 86.6 & 12.8 & 25.5 & 28.0 & 56.6 & 92.8 & 13.4 \\
  \midrule
  \midrule
  \multicolumn{6}{l}{  \it{Models trained with CIDEr optimization} } \\
     \midrule
  \midrule
  \ModelName-B &  34.2 & 30.9 & 60.0 & 108.9 & 14.0 & 28.8 & 28.4 & 58.2 & 96.0 & 13.5 & 19.8 & 26.0 & 55.6 & 90.1 & 12.7 & 27.7 & 28.3 & 57.9 & 96.7 & 13.4   \\
  \ModelName-GOLD-B & 35.4 & 31.2 & 60.6 & 110.3 & 14.3 & 30.5 & 29.0 & 58.9 & 99.5 & 13.8 & 20.4 & 26.0 & 55.6 & 90.2 & 12.8 & 29.1 & 28.7 & 58.5 & 99.2 & 13.7\\
  \midrule
  \end{tabular}
  }
  \vspace{-2mm}
  \caption{The complete set of results on the NoCaps validation set. B@4: BLEU@4, M: METEOR, R: ROUGE-L, C: CIDEr, S: SPICE.
  }
  \label{tab:caption_full_nocaps_val}
   \end{center}
\end{table*}

\begin{table*}[t]
\setlength{\tabcolsep}{3pt}
\begin{center}
\resizebox{\textwidth}{!}
{
 \begin{tabular}{cccccccccccccccccccccccccccccccccccccc} 
 \toprule
    \multirow{2}{*}{\bf Model}  & \multicolumn{5}{c}{\bf in-domain} & \multicolumn{5}{c}{\bf near-domain} & \multicolumn{5}{c}{\bf out-domain} & \multicolumn{5}{c}{\bf entire}  \\
    \cmidrule(lr){2-6}    \cmidrule(lr){7-11}    \cmidrule(lr){12-16}    \cmidrule(lr){17-21} 
     & \bf B@4 & \bf M & \bf R & \bf C &\bf S   & \bf B@4 & \bf M & \bf R & \bf C &\bf S   & \bf B@4 & \bf M & \bf R & \bf C &\bf S  & \bf B@4 & \bf M & \bf R & \bf C &\bf S  \\
    \midrule
     \multicolumn{6}{l}{  \it{Models trained without CIDEr optimization} } \\
     \midrule
  \ModelName-B &  28.6 & 29.5 & 57.7 & 92.8 & 13.6 & 25.6 & 28.0 & 56.1 & 87.3 & 13.0 & 16.2 & 24.4 & 52.1 & 76.4 & 11.6 & 24.3 & 27.6 & 55.6 & 86.0 & 12.9 \\ 
  \ModelName-GOLD-B & 29.9 & 30.1 & 58.4 & 95.9 & 14.1 & 28.0 & 28.7 & 57.4 & 92.0 & 13.5 & 18.4 & 25.3 & 53.4 & 81.0 & 12.3 & 26.5 & 28.3 & 56.8 & 90.6 & 13.4 \\ 
  \midrule
  \midrule
  \multicolumn{6}{l}{  \it{Models trained with CIDEr optimization} } \\
     \midrule
  \midrule
  \ModelName-B & 33.3 & 30.4 & 59.9 & 102.7 & 14.1 & 29.8 & 28.9 & 58.6 & 95.3 & 13.6 & 20.7 & 25.5 & 55.1 & 83.4 & 12.3 & 28.6 & 28.5 & 58.2 & 94.1 & 13.4 \\
  \ModelName-GOLD-B &     34.6 & 30.9 & 60.6 & 104.7 & 14.4 & 31.3 & 29.4 & 59.4 & 98.7 & 13.9 & 21.2 & 25.9 & 55.4 & 85.7 & 12.7 & 29.9 & 29.0 & 58.8 & 97.1 & 13.8 \\
  \midrule
  \end{tabular}
  }
  \vspace{-2mm}
  \caption{The complete set of results on the NoCaps test set. B@4: BLEU@4, M: METEOR, R: ROUGE-L, C: CIDEr, S: SPICE.
  }
  \label{tab:caption_full_nocaps_test}
   \end{center}
\end{table*}

%% file: supp_tables/open_vqa.tex
\begin{wraptable}{R}{0.5\textwidth}
\vspace{-4mm}
\setlength{\tabcolsep}{5pt}
\begin{center}
  \begin{tabular}{cccccccccc}
  \toprule
    \multirow{2}{*}{\bf Model}  &  \multicolumn{3}{c}{\bf Open-ended VQA}  \\
    \cmidrule(lr){2-4}  
   &  \bf In-D & \bf Out-D & \bf overall  \\
    \midrule
VL-T5~\cite{cho2021unifying} & 71.4 & 13.1 & 67.9 \\
VL-BART~\cite{cho2021unifying} & 72.1 & 13.2 & 68.6 \\
\demph{SimVLM-B~\cite{wang2021simvlm}} & \demph{78.3} & \demph{25.8} & \demph{75.2} \\
\midrule
\ModelName-B & 75.9 & 14.7 & 71.6 \\
    \midrule
  \end{tabular}
  \caption{Results on open-ended VQA. We follow~\cite{cho2021unifying} to split the data into in domain (In-D) and out of domain (out-D).}
  \label{tab:open_vqa}
     \end{center}
   \vspace{-3mm}
\end{wraptable}

%% file: supp_tables/uni_modal.tex
\begin{table}[t!]
\small
\begin{center}
  \begin{tabular}{cccccccccc}
    \toprule
    \multirow{1}{*}{\bf Text Encoder} & \bf QQP & \bf  MNLI & \bf QNLI & \bf SST2 & \bf CoLA & \bf MRPC & \bf STSB & \bf RTE  \\
    \midrule
  RoBERTa-B~\cite{liu2019roberta} & 91.31  & 87.53  & 92.61  & 94.38 & 58.72  & 91.03  & 90.15 & 71.24  \\
  METER-RoBERTa-B~\cite{dou2021empirical} & 91.34  & 87.38  & 92.67 & 93.92  & 57.88  & 90.57  & 89.93 & 70.28   \\
  SimVLM-B~\cite{wang2021simvlm} & 90.4 & 83.4 & 88.6  & 90.9 & 46.7 & 84.4 & - & 63.9  \\
  \midrule
  \ModelName-RoBERTa-B & 91.60  & 86.23 & 91.34 & 92.66 & 59.56 & 90.72 & 89.77 & 62.09 \\
  \midrule
  \end{tabular}
  \caption{Performance of text encoders on the GLUE dev sets.}
  \label{tab:glue}
       \end{center}
   \vspace{-3mm}
\end{table}

\begin{table}[ht!]
\small
\begin{center}
\begin{tabular}{cccccccccc}
     \toprule
     \multirow{1}{*}{\bf Image Encoder} & \bf ImageNet & \bf ADE20k  \\
     \midrule
   Swin-B & 86.3 & 51.6 \\
   \midrule
   \ModelName-Swin-B & 86.0 & 52.0 \\
   \midrule
   \end{tabular}
   \caption{Performance of image encoders on image classification and semantic segmentation.}
   \label{tab:image_encoder}
       \end{center}
   \vspace{-3mm}
 \end{table}

%% file: supp_tables/ODinW.tex
\begin{table}[t]
\setlength{\tabcolsep}{2pt}
\begin{center}
\tiny
\def \arraystretch{0.98}
\begin{tabular}{cccccccccccccccc}
 \toprule
\bf Model & \bf Shot & \bf PascalVOC & \bf AerialDrone & \bf Aquarium & \bf Rabbits &
\bf EgoHands & \bf Mushrooms &
\bf Packages &\bf  Raccoon & \bf Shellfish &\bf  Vehicles & \bf Pistols & \bf Pothole & \bf Thermal & \bf Avg \\
 \midrule
GLIP-B & 1 & 51.7 & 25.0 & 34.0 & 69.2 & 67.4 & 63.4. & 54.4  & 56.5 & 14.1 &57.9 & 51.7 & 15.6 & 70.2 & 48.5\\
GLIP-B & 3 & 57.4 & 25.2 & 44.9 & 65.1 & 69.3 & 88.3 & 67.3 & 52.8 & 28.9 & 60.7 & 62.1 & 31.3 & 67.9 & 55.5 \\
GLIP-B & 5 & 57.9  & 28.1 & 44.1 & 64.6 & 68.1 & 85.1 & 74.2 & 60.8 & 24.0 & 61.9 & 59.1 & 33.7 & 75.2 & 56.7 \\
GLIP-B & 10  & 60.6 & 27.8 & 49.7 & 67.8 & 65.2 & 87.4 & 67.5 & 54.5 & 42.3 & 65.1 & 63.9 & 39.2 & 78.7 & 59.2 \\
GLIP-B & All  & 62.2  & 36.0 &55.3 & 74.0 & 79.8 & 88.1 & 74.3& 64.1 & 47.0 & 64.4 & 72.4 & 56.5 & 81.1 & 65.8 \\
  \midrule
\ModelName-B & 1 & 55.7 & 25.0 & 37.9 & 69.8 & 67.2 & 83.0 & 73.2  & 54.7 & 29.7 & 58.0 & 44.8 & 27.8 & 67.0 & 53.4 \\
\ModelName-B & 3 & 59.8 & 28.2 & 42.9 & 71.5 & 68.4 & 88.1 & 65.6  & 64.6 & 38.6 & 61.8 & 47.8 & 37.5 & 68.5 & 57.2 \\
\ModelName-B & 5 & 61.6 & 30.9 & 49.5 & 72.3 & 69.2 & 87.5 & 73.2  & 57.4 & 38.2 & 62.7 & 55.3 & 40.3 & 61.8 & 58.4  \\
\ModelName-B & 10 & 60.7  & 31.2 & 52.0 & 68.8 & 70.4 & 88.1 & 69.3 & 53.8 & 41.7 & 66.6 & 62.2 & 46.8 & 74.8 & 60.5\\
\ModelName-B & All  & 68.7 & 35.4 & 58.3 & 75.9 & 79.4 & 88.1 & 72.2 & 52.9 & 45.2 & 66.1 & 72.2 & 60.9 & 80.6 & 65.9\\
\midrule
\end{tabular}
\caption{ Few-shot and full fine-tuned results on the various ODinW datasets.
\label{tab:odinw_fullres}}
\end{center}
\vspace{-3mm}
\end{table}